\documentclass{article} %
\usepackage{iclr2027_conference,times}

\usepackage{amsmath,amsfonts,bm}

\usepackage{bbold}

\def\eqref#1{equation~\ref{#1}}

\def\1{\bm{1}}

\def\vk{{\bm{k}}}

\def\vo{{\bm{o}}}

\def\vq{{\bm{q}}}

\def\vu{{\bm{u}}}
\def\vv{{\bm{v}}}

\def\vx{{\bm{x}}}

\def\mB{{\bm{B}}}
\def\mC{{\bm{C}}}
\def\mD{{\bm{D}}}

\def\mI{{\bm{I}}}

\def\mK{{\bm{K}}}

\def\mM{{\bm{M}}}

\def\mO{{\bm{O}}}

\def\mQ{{\bm{Q}}}

\def\mS{{\bm{S}}}

\def\mV{{\bm{V}}}
\def\mW{{\bm{W}}}

\DeclareMathAlphabet{\mathsfit}{\encodingdefault}{\sfdefault}{m}{sl}
\SetMathAlphabet{\mathsfit}{bold}{\encodingdefault}{\sfdefault}{bx}{n}

\newcommand{\softmax}{\mathrm{softmax}}

\usepackage[dvipsnames]{xcolor}
\definecolor{linkblue}{RGB}{70,160,210}
\usepackage[colorlinks=true,linkcolor=linkblue,citecolor=linkblue,urlcolor=linkblue]{hyperref}

\usepackage{graphicx}
\usepackage{subcaption}
\usepackage{tikz}
\usetikzlibrary{positioning}
\usepackage{tabularx}
\usepackage{multirow}
\usepackage{wrapfig}
\usepackage{tabularx}
\usepackage{booktabs}
\usepackage{xltabular}   %
\usepackage{wrapfig}
\usepackage{graphicx}
\usepackage{colortbl}
\usepackage{caption}
\usepackage{comment}

\usepackage{amsmath}
\usepackage{amssymb}
\usepackage{mathtools}
\usepackage{amsthm}

\usepackage{algorithm}
\usepackage{algpseudocode}

\usepackage[framemethod=TikZ]{mdframed}
\usepackage{tcolorbox}
\definecolor{chart}{HTML}{1f77b4}
\definecolor{goldrow}{HTML}{E7E6F7}   %
\definecolor{hlcell}{HTML}{FCF3B8}    %
\definecolor{qarow}{HTML}{ECECEC}     %
\mdfsetup{%
    middlelinecolor =   none,
    middlelinewidth =   1pt,
    backgroundcolor =   blue!5,
    roundcorner     =   5pt,
}
\newtcolorbox{example}[1][]{
  colback=chart!5!white,
  colframe=chart,
  floatplacement=floating,
  title=\centering \textsf{#1}
}

\usepackage{xspace}

\newcommand{\ie}{\textit{i.e.}\xspace}
\newcommand{\merlin}{\textsc{Merlin}\xspace}
\newcommand{\ours}{\merlin}
\newcommand{\oursfull}{\textbf{M}om\textbf{E}nt \textbf{R}outing for multi-state \textbf{LIN}ear attention\xspace}

\newcommand{\seqlen}{T}                       %
\newcommand{\hdim}{d}                         %
\newcommand{\rdim}{d_r}                       %
\newcommand{\nseg}{L}                         %
\newcommand{\seg}{C_i}                        %
\newcommand{\segsize}{m}                      %
\newcommand{\segstate}{\mS_{C_i}}             %
\newcommand{\decay}{\alpha_t}                 %
\newcommand{\gate}{\gamma_t}                  %
\newcommand{\aggw}{\lambda_{t,i}}             %
\newcommand{\rquery}{\vq'_t}                  %
\newcommand{\rkey}{\vk'_t}                    %
\newcommand{\firstm}{\bm{\mu_i}}                   %
\newcommand{\secondm}{\bm{\Sigma_i}}                  %
\newcommand{\rmean}{r^\mu_{t,i}}                %
\newcommand{\rvar}{r^\Sigma_{t,i}}                 %
\newcommand{\taum}{\tau_\mu}                    %
\newcommand{\tauv}{\tau_\Sigma}                    %
\newcommand{\gsmu}{\eta_{t,i}}                 %
\newcommand{\gsvar}{\sigma^2_{t,i}}           %
\newcommand{\gsA}{A_{t,i}}                    %
\newcommand{\gsZ}{Z_{t,i}}                    %

\usepackage{titletoc}

\theoremstyle{plain}

\theoremstyle{definition}

\theoremstyle{remark}

\usepackage[capitalize]{cleveref}
\Crefname{table}{Tab.}{Tabs.}
\crefname{lemma}{lemma}{lemmas}
\Crefname{lemma}{Lemma}{Lemmas}
\crefname{proposition}{proposition}{propositions}
\Crefname{proposition}{Proposition}{Propositions}
\crefname{theorem}{theorem}{theorems}
\Crefname{theorem}{Theorem}{Theorems}
\crefname{corollary}{corollary}{corollaries}
\Crefname{corollary}{Corollary}{Corollaries}
\crefname{assumption}{assumption}{assumptions}
\Crefname{assumption}{Assumption}{Assumptions}
\crefname{remark}{remark}{remarks}
\Crefname{remark}{Remark}{Remarks}

\usepackage[disable,textsize=tiny]{todonotes}

\usepackage{url}

\title{Multi-State Linear Attention with \\Generalized Second-Order Routing}

\author{
    Hee Seung Hwang$^{1}$
    \quad
    Sanghyuk Chun$^{1}$
    \quad
    Xindi Wu$^{1}$
    \quad
    Olga Russakovsky$^{1}$ 
    \quad
    Zhiwei Deng$^{2}$ \\ \\
    $^{1}$Princeton University \quad $^{2}$Google DeepMind
}

\iclrpreprint %
\begin{document}

\maketitle
\hypertarget{abstract}{}

\makeatletter
\let\BD@acl\addcontentsline
\renewcommand{\addcontentsline}[3]{}
\makeatother

\begin{abstract}

Self-attention plays a key role in long-sequence modeling, but its quadratic time complexity fundamentally limits the scaling of Large Language Models. Recently, Multi-State Linear Attention (MSLA) mechanisms have been proposed to make sequence modeling more efficient by splitting the context into segments and compressing each in its own memory state. MSLA aggregates the information extracted from each memory state using only the mean alignment between a query and each segment's keys, ignoring how the alignment is distributed within the segment. In order to model the relevance of a segment to a query more comprehensively and
incorporate the variance of the alignment in the aggregation,
we propose \ours (\oursfull), a method that tracks the first and second moments of each segment's keys. Using the moments of keys,
a query can recover both the mean and the variance of its alignment with each segment to perform a more content-aware aggregation of per-state readouts. 
Experiments on multi-state variants of Gated DeltaNet-1.3B and Mamba-2-780M show that \ours consistently outperforms existing aggregation methods on various long-context tasks. 
In particular, \ours outperforms baselines on RULER needle-in-a-haystack and long-context QA (+8.3\% for Gated DeltaNet-1.3B and +7.0\% for Mamba-2-780M) and diverse subtasks in LongBench (+13.5\% and +12.0\%). It even outperforms baselines that consume 4x more memory. \ours provides a simple and general routing mechanism for improving long-context capabilities of MSLA models.

\end{abstract}

\section{Introduction}

The self-attention mechanism is a core operation in long-sequence modeling \citep{vaswani2017attention}. However, its quadratic computational complexity and linearly growing key-value (KV) cache have become a key bottleneck in scaling compute and data \citep{keles2023computational}. Linear attention \citep{katharopoulos2020transformers,sun2023retentive,peng2023rwkv} addresses these limitations by reformulating attention to operate in sub-quadratic complexity. For example, DeltaNet \citep{yang2024parallelizing} and Gated DeltaNet \citep{yang2025gated} remove the softmax in standard attention and replace the KV cache with a single fixed-size memory state. However, a single fixed-size memory state is inherently limited in memory capacity and may lead to information loss when compressing longer contexts \citep{arora2024simple,jelassi2024repeat, kuratov2024babilong}. As a result, models based on single-state linear attention still underperform standard attention in practice.

\begin{figure}[t]
\centering
\begin{subfigure}[b]{0.68\textwidth}
\centering
\includegraphics[width=\linewidth]{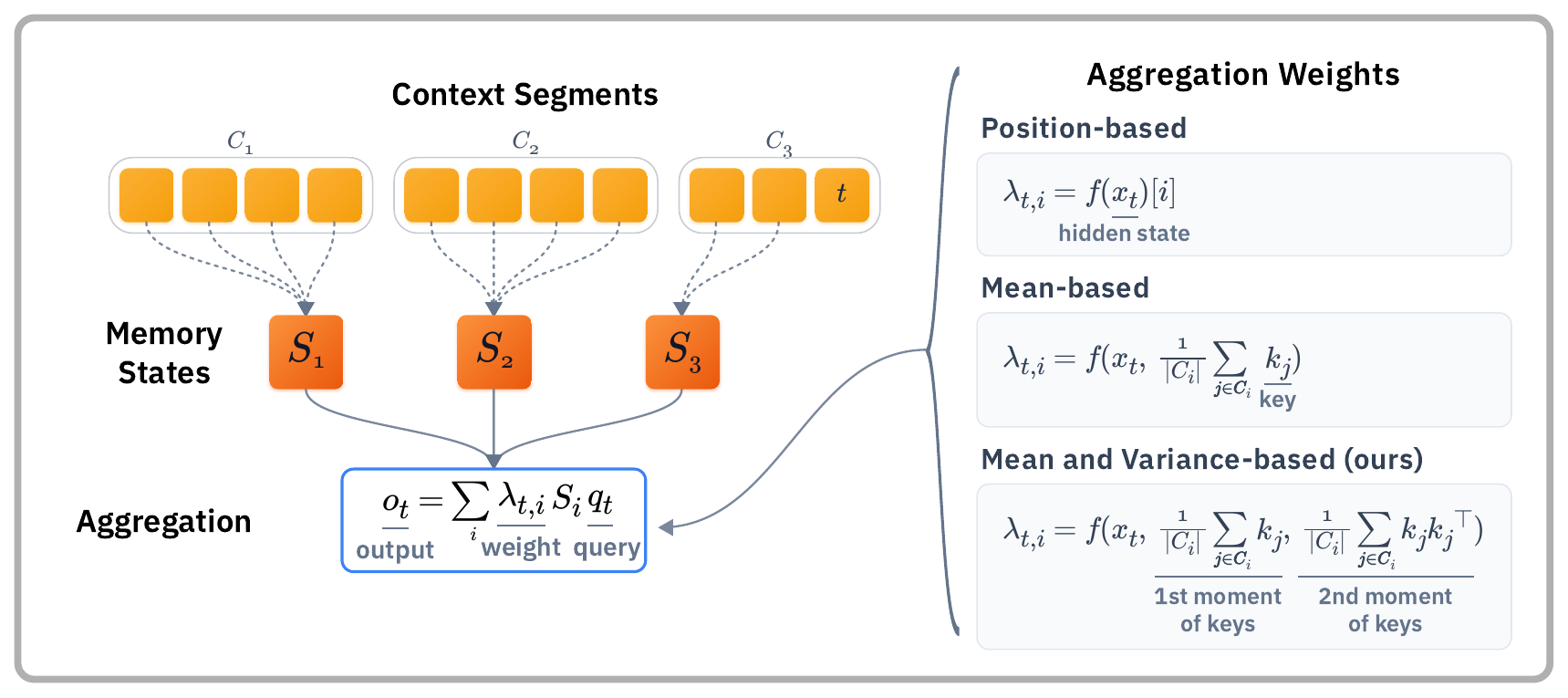}
\caption{Multi-State Linear Attention (MSLA)}
\label{fig:banner_arch}
\end{subfigure}\hfill
\begin{subfigure}[b]{0.30\textwidth}
\centering
\includegraphics[width=\linewidth]{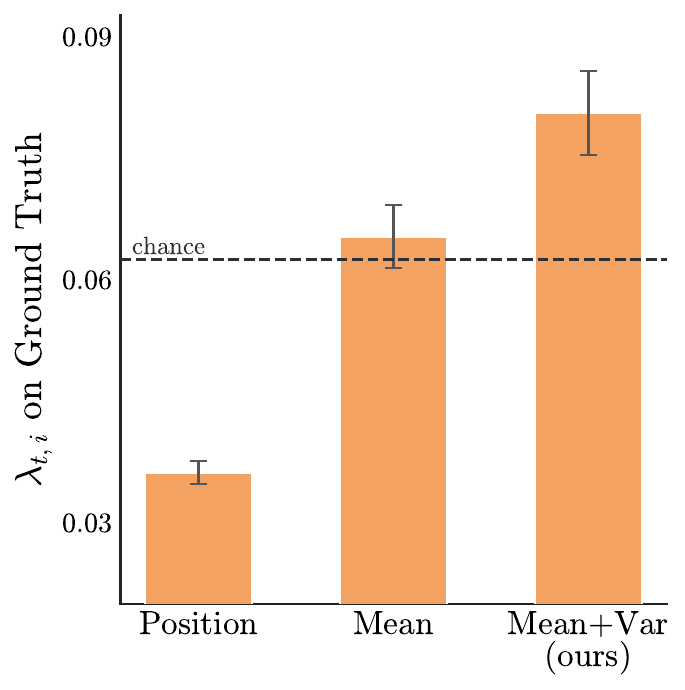}
\caption{Aggregation Weight on GT}
\label{fig:banner_bar}
\end{subfigure}
\caption{\small \textbf{(a)} A multi-state linear attention (MSLA) model aggregates segment-level outputs via a weighted sum. Prior methods compute the weights $\aggw$ from position or from the mean of each segment's keys. \ours instead reads both the first and second moments of the keys.
\textbf{(b)} On retrieval-augmented generation samples in HELMET~\citep{yen2025helmet}, we measure the aggregation weight each method places on the ground truth (GT) segment at the query token. Position-based aggregation largely misses the GT and mean-based aggregation recovers more, while \ours places the most weight on it.
}
\vspace{-10pt}
\label{fig:methods_banner}
\end{figure}

Recently, Multi-State Linear Attention (MSLA) mechanisms \citep{behrouz2026memory, guo2026log, wang2026dynamic} have been introduced to improve the long-context capabilities of single-state linear attention. MSLA partitions the context into segments and compresses each segment into its own memory state. Each token extracts segment-level outputs from preceding memory states and then aggregates them to produce its final output representation (\cref{fig:banner_arch}). Operating in the multi-state regime therefore introduces two novel challenges: how to divide the context into segments (\textit{\textbf{separation}}), and how to combine the segment-level outputs (\textit{\textbf{aggregation}}). 

We study \textit{aggregation}, which is relatively underexplored compared to \textit{separation}. Existing methods aggregate segment-level outputs via a weighted sum \citep{behrouz2026memory, guo2026log, wang2026dynamic}, where the weights represent each segment's relevance to the query: closely aligned segments contribute more to the final output than others. However, prior works derive the weights either from position \citep{guo2026log,wang2026dynamic}, assuming recency as the source of relevance, or from the mean (first-order) alignment between each segment's keys and the query \citep{behrouz2026memory}. They overlook the variance (second-order) of the alignment, which captures sharply aligned keys that the mean alone cannot distinguish from a segment of uniformly weak ones.

To analyze why the variance matters, we examine how an MSLA model assigns aggregation weights on a long-sequence reasoning task. \cref{fig:banner_bar} shows that the segment containing the ground truth, which is the segment most relevant to the response, receives the highest aggregation weight when the weights are derived from both the mean and variance of the alignment, compared to the mean or position alone. This motivates an aggregation method that incorporates the variance of the alignment in deriving the weights.

We propose \ours (\oursfull), a method that leverages the mean and variance of key-query alignment to improve aggregation in multi-state linear attention. \ours stores the first and second moments of each segment's keys. Using the moments, a query recovers both the mean (first-order) and variance (second-order) of its alignment with each segment's keys to derive the weights for the weighted sum of segment-level outputs. Our experiments show that \ours improves the long-context capabilities of multi-state linear attention models.

\textbf{Our contributions are:}
\begin{enumerate}
\item We identify segment-level output aggregation as a key bottleneck in multi-state linear attention: existing methods weight each segment's output using only the mean alignment between the query and the segment's keys (\S\ref{sec:preliminary}).
\item We propose \ours, a method that keeps the first and second moments of each segment's keys. A query uses these moments to recover the mean and variance of its alignment with the segment (\S\ref{sec:key_moment}) and to derive the aggregation weights (\S\ref{sec:weight_derivation}).
\item We evaluate multi-state variants of Gated DeltaNet-1.3B and Mamba-2-780M (\S\ref{sec:setup}). On LongBench~\citep{bai2024longbench}, \ours improves the average score over the strongest baseline by 13.5\% on Gated DeltaNet and 12.0\% on Mamba-2 (\S\ref{main_results}).
\end{enumerate}
\vspace{-10pt}

\section{Background}
\label{sec:preliminary}

Given an input sequence of $\seqlen$ tokens, the query, key, and value matrices $\mQ, \mK, \mV \in \mathbb{R}^{\seqlen \times \hdim}$ are projected from the tokens' hidden states $\vx_1, \dots, \vx_{\seqlen}$. Standard full attention computes:
$$\text{\textcolor{gray}{\textbf{Attention} \citep{vaswani2017attention}:}}\quad \mO = \textcolor{magenta}{\softmax}(\mQ \mK^\top +  \mM)\mV,$$
where $\mM \in \{-\infty, 0\}^{T \times T}$ is a causal masking matrix.
Softmax attention costs quadratic time, and requires a KV cache that grows linearly with the context. This motivates more efficient sub-quadratic attention formulations.

\textbf{Linear Attention.} Linear attention removes the softmax normalization and instead computes:
$$\text{\textcolor{gray}{\textbf{Linear Attention}:}}\quad \mO = (\mQ \mK^\top \odot \mM)\mV, \quad \mM_{i,j} = 1 \text{ if } i \geq j, \text{ and } 0 \text{ otherwise}.$$
This formulation can be reorganized into a recurrence over a memory state $\mS_t \in \mathbb{R}^{\hdim \times \hdim}$:
\begin{equation}
\label{eq:linear_attention}
\text{\textcolor{gray}{\citet{katharopoulos2020transformers}:}}\quad\mS_t = \mS_{t-1} + \vv_t \vk_t^\top = \sum_{t'} \vv_{t'} \vk_{t'}^\top, \quad \vo_t = \mS_t \vq_t, 
\end{equation}
where $\vk_t, \vv_t, \vq_t$ are the key, value, and query vectors of a token at position $t$. The memory state $\mS_t$ accumulates key-value outer products, acting as an associative memory that binds each key to its value. A query reads from the memory state via a matrix-vector product to get the output representation $\vo_t$. The recurrence now runs in linear time and constant memory. Recent works add gating and delta-rule correction to improve the memory capacity of the fixed-size state. For example, Gated DeltaNet~\citep{yang2025gated} (GDN) uses the recurrence as follows:
\begin{equation}
\text{\textcolor{gray}{\textbf{Gated DeltaNet} \citep{yang2025gated}:}}\quad    \mS_t = \mS_{t-1}\big(\decay(\mI - \beta_t \vk_t \vk_t^\top)\big) + \beta_t \vv_t \vk_t^\top,
    \label{eq:gdn}
\end{equation}
where $\decay \in (0,1)$ decays the state and $\beta_t$ sets the strength of the update.

\textbf{Multi-State Linear Attention.} 
Linear attention can be limiting when $\seqlen$ grows because it needs to compress the entire context into a single fixed-size state. 
Multi-state linear attention relaxes this bottleneck by partitioning the sequence into $\nseg$ contiguous segments $C_1, \dots, C_L$ and compressing each segment in its own state:
\begin{equation}
\text{\textcolor{gray}{\textbf{Multi-State Linear Attention (Separation)}:}}\quad    \segstate = \sum_{j \in \seg} \vv_j \vk_j^\top.
    \label{eq:separation}
\end{equation}
Intuitively, \cref{eq:separation} can be viewed as splitting the first equation of \cref{eq:linear_attention} into $L$ parts, which we call \textit{separation}.
At step $t$, a query reads segment-level outputs from the last $\ell \leq L$ memory states visible at step $t$. The intermediate outputs are then aggregated via a weighted sum to derive the final output representation:
\begin{equation}
\text{\textcolor{gray}{\textbf{Multi-State Linear Attention (Aggregation)}:}}\quad    \vo_{t, i} = \mS_{C_i} \vq_t,  
    \quad 
    \vo_t = \sum_{i=1}^{\ell} \aggw\, \vo_{t, i},
    \label{eq:aggregation}
\end{equation}
where $\aggw$ is the aggregation weight that determines how much the output from segment $i$ contributes to the final output at position $t$.
\cref{eq:aggregation} can be viewed as the second equation of \cref{eq:linear_attention} applied to $\ell$ partitioned states, which we refer to as aggregation.

Unlike single-state linear attention, in multi-state linear attention, the number of memory states increases as the content length increases. Multi-state linear attention therefore interpolates between single-state linear attention (constant memory, linear time) and full attention (linear memory, quadratic time)~\citep{behrouz2026memory}. \cref{tab:complexity} summarizes the time and memory complexities of attention variants.

\begin{table}[t]
\centering
\footnotesize
\caption{\small \textbf{Computational and Memory Complexity.}  \emph{Memory} is the size of the state(s) retained to decode each token. $\rdim$ is the dimension of the moments of keys (\S\ref{sec:key_moment}).  Memory Caching~\citep{behrouz2026memory} and \ours assume uniform segmentation ($\nseg=\seqlen/m$) where $m$ is the number of tokens per segment.
}
\vspace{-5pt}
\label{tab:complexity}
\begin{tabular}{l c c}
\toprule
\textbf{Mechanism} & \textbf{Compute (Prefill)} & \textbf{Memory} \\
\midrule
Full Attention~\citep{vaswani2017attention}        & $O(\seqlen^2 \hdim)$              & $O(\seqlen \hdim)$ \\
Linear Attention~\citep{katharopoulos2020transformers}                & $O(\seqlen \hdim^2)$             & $O(\hdim^2)$ \\
\textit{\textcolor{gray}{Multi-State Linear Attention}}    & & \\
\quad Log-Linear Attention~\citep{guo2026log}      & $O(\seqlen \hdim^2 \log \seqlen)$ & $O(\hdim^2 \log \seqlen)$ \\
\quad Memory Caching~\citep{behrouz2026memory}            & $O(\seqlen \nseg \hdim^2)$       & $O(\nseg \hdim^2)$ \\
\quad \ours{} (ours)            & $O(\seqlen \nseg (\hdim^2+\rdim^2))$ & $O(\nseg (\hdim^2+\rdim^2))$ \\
\bottomrule
\end{tabular}
\vspace{-15pt}
\end{table}

\textbf{Aggregation in Multi-State Linear Attention.} The aggregation of segment-level outputs in \cref{eq:aggregation} requires weights $\aggw$. These weights function as routers that increase the contribution of outputs from query-relevant segments and decrease the contribution of outputs from others. The simplest methods learn a position-based weight vector $\boldsymbol{\lambda}_t$ (with entries $\aggw$) directly from the target token, projected from its query or hidden state~\citep{guo2026log,wang2026dynamic}:
\begin{equation}
    \text{\citet{guo2026log}:}\quad \boldsymbol{\lambda}_{t} = \mW_{\lambda} \vx_t \quad \text{or} \quad \text{\citet{wang2026dynamic}:} \quad \boldsymbol{\lambda}_{t} = \mW_{\lambda} \vq_t.
    \label{eq:position}
\end{equation}
These weights do not consider what each segment stores and therefore cannot function as a content-aware router. \citet{behrouz2026memory} introduces content-awareness in aggregation weights as follows: 
\begin{equation}
    \text{\citet{behrouz2026memory}:}\quad \aggw = {\langle \bar{\vk}_i, \vu_t \rangle}, 
    \qquad \vu_t = \mW_u \vx_t , 
    \qquad \bar{\vk}_i = \frac{1}{|\seg| }\sum_{j \in \seg} \vk_j,
    \label{eq:mc}
\end{equation}
where $\vu_t$ is a learned vector projected from hidden state $\vx_t$ and $\bar{\vk}_i$ is the first moment of keys in segment $i$ (\ie, the mean). However, this only captures the \textit{mean} alignment between the query and a segment's keys and does not capture the \textit{variance}. 

We revisit the analysis in \cref{fig:banner_bar} and demonstrate that the variance of the alignment is necessary for more content-aware aggregation of segment-level readouts. We train multi-state models with different aggregation methods and evaluate on 30 randomly selected TriviaQA samples in HELMET~\citep{yen2025helmet} at 4K context length. This is a retrieval-augmented generation task where each sample consists of multiple passages and a prompt that requires retrieving information from one of them. Thus, the model should locate the relevant passage and extract the correct answer to the prompt. 

We find that different aggregation methods assign varying amounts of aggregation weight to the segment containing the ground truth. Ideally, the weights assigned to the ground truth segment should be higher than others in order to increase the contribution of the information extracted from the segment in the final response. \cref{fig:banner_bar} shows that mean- and variance-based aggregation assigns the highest weight to the ground truth segment compared to position-only and mean-only aggregation. This motivates our proposed method \ours, which leverages both orders of alignment to compute the weights for segment-level output aggregation.

\section{\ours}
\label{sec:method}

\begin{figure}[t]
\centering
\includegraphics[width=\textwidth]{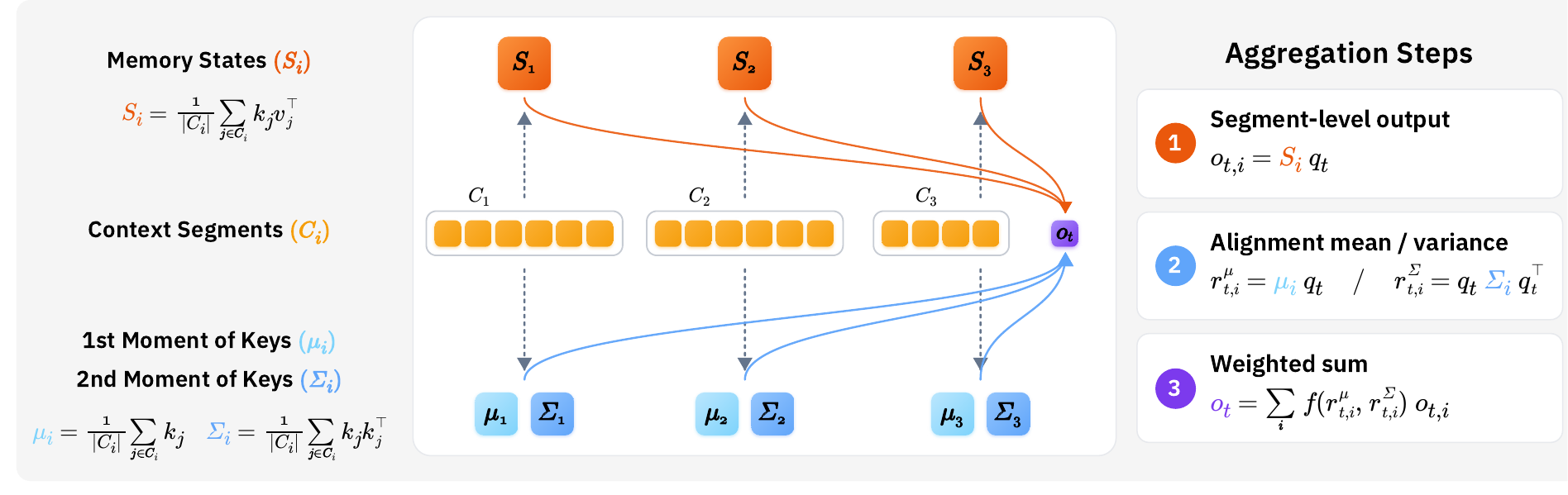}
\vspace{-10pt}
\caption{\textbf{\ours Overview.} The context is split into segments $\seg$ (middle row). Each segment stores a memory state (top row) and the first and second moments of its keys (bottom row). At position $t$, the query $\vq_t$ reads segment-level outputs from memory states (step 1). The query computes the mean and variance of the key-query alignment from the moments (step 2). The final output is a weighted sum of segment-level outputs, where the weights are derived from a gated mix of the alignment mean and variance (step 3). 
}
\label{fig:gram}
\end{figure}

Our goal is to capture the variance of the alignment between a query and each segment's keys to derive more informative aggregation weights. %
We first introduce how the first and second moments of the keys are constructed (\S\ref{sec:key_moment}). Then, we discuss how the mean and variance of the alignment can be derived from the moments (\S\ref{sec:weight_derivation}). Finally, we elaborate on the relationship 
between multi-state linear attention equipped with \ours and softmax attention (\S\ref{sec:connection}).

\subsection{First and Second Moments of Keys}
\label{sec:key_moment}

For each token, we project new keys and queries $\vk_t', \vq_t' \in \mathbb{R}^{\rdim}$ from the hidden states. We use $\vk_t', \vq_t'$ instead of $\vk_t, \vq_t$ to construct and read from the first and second moments of keys, primarily to keep moments compact by choosing $\rdim$ that is smaller than $d$. 

The first moment of keys is the mean-pooled vector of all keys ($\vk_j'$) in the segment. The second moment is the average of the outer products $\vk_j'\vk_j'^\top$ over the segment. The first and second moments of keys can thus be formulated as:  
\begin{equation}
    \firstm = \frac{1}{|\seg|} \sum_{j\in \seg} \vk_j', \quad
    \secondm = \frac{1}{|\seg|}\sum_{j\in \seg} \vk_j' \vk_j'^\top.
    \label{eq:moments}
\end{equation}
The first and second moments are fixed-size, and their dimensions depend only on $\rdim$. Together, they summarize segment $i$, where $\firstm$ captures the average direction of the keys and $\secondm$ captures how the keys are spread within the segment.  

\subsection{Aggregation Weight Derivation}
\label{sec:weight_derivation}

Given a token at position $t$, we compute the following scalar scores for each visible segment $\seg$ (the segments leading up to the current one):
\begin{equation}
    {\rmean} = \firstm^\top \rquery,
\quad
\rvar = {\rquery}^\top \secondm \rquery - (\rmean)^2.
    \label{eq:score}
\end{equation}
Here, $\rmean$ captures the mean key-query alignment, measuring how well the segment's keys broadly align with the query. Meanwhile, $\rvar$ captures the variance, measuring the spread of the per-token alignment between the query and the keys in the segment. 
We mix the softmax-normalized $\rmean$ and $\rvar$ with a learned gate to get the final weight $\aggw$: %
\begin{equation}
\text{\textbf{\ours (ours):}}\quad
\aggw = \gate \softmax_i(\taum\, \rmean) + (1 - \gate) \softmax_i(\tauv\, \rvar),
\label{eq:weight}
\end{equation}
where $\softmax_i$ denotes the softmax operation over the segment index $i$, $\gate$ is a scalar gate, and $\taum$ and $\tauv$ are learnable temperatures that scale the two scores before normalization.
The gate $\gate$ is computed from the token's hidden state $\vx_t$ through a learned projection. The temperatures $\taum, \tauv$ calibrate the sharpness of each term. This shows how the first and second moments are used to derive the weights for \cref{eq:aggregation}.

\subsection{Connection to Softmax Attention}
\label{sec:connection}

Suppose we ran token-level softmax attention over the keys and then grouped its mass by segment. For a target token at position $t$ and a segment $\seg$, the key-query similarity can be expressed as:
$$
s_{t,i,j} = {\rquery}^{\top} \vk'_j, \qquad j \in \seg.
$$
We can define the length-normalized evidence of the segment as:
$$
\gsZ(\delta) = \frac{1}{|\seg|} \sum_{j \in \seg} e^{\delta\, s_{t,i,j}},
$$
where $\delta$ is a temperature hyperparameter. Grouping the token-level softmax by segment, the total attention mass a full softmax layer would place on segment $i$ can be expressed as:
\begin{equation}
\gsA
=
\frac{
    \sum_{j \in \seg} \exp\!\left(\delta s_{t,i,j}\right)
}{
    \sum_{\ell} \sum_{j \in C_\ell}
    \exp\!\left(\delta s_{t,\ell,j}\right)
}
=
\operatorname{softmax}_{i}
\left(
    \log |\seg| + \log \gsZ(\delta)
\right).
\end{equation}
Because our segments are equal-sized, $\log |\seg|$ is the same for every segment and cancels inside the softmax. Expanding the remaining log-evidence in cumulants around $\delta = 0$, we obtain:
\begin{equation}
\log \gsZ(\delta) = \delta\, \gsmu + \tfrac{\delta^{2}}{2}\, \gsvar + O(\delta^{3}), 
\quad 
\gsmu = \frac{1}{\segsize} \sum_{j\in \seg} s_{t,i,j},
\quad 
\gsvar = \frac{1}{\segsize} \sum_{j\in \seg} s_{t,i,j}^{2} - \gsmu^{2}
\label{eq:gsa-cumulant}
\end{equation}
Therefore, the relevance of a segment is governed by the first two terms: the mean $\gsmu$ and the variance $\gsvar$. These are precisely the scores $\rmean$ and $\rvar$, which are computed from the first and second moments via matrix-vector products:
\begin{equation}
\gsmu = \firstm^{\top} \rquery = \rmean, \qquad
\gsvar = {\rquery}^{\top} \secondm\, \rquery - \bigl(\firstm^{\top} \rquery\bigr)^{2} = \rvar.
\label{eq:gsa-recover}
\end{equation}
Rather than the fixed additive coefficients of \cref{eq:gsa-cumulant} ($\delta$ on the mean, $\tfrac{\delta^{2}}{2}$ on the variance) inside a single softmax, \ours combines the two scores with a learned gated mixture of separately-normalized softmax terms (\cref{eq:weight}), with learnable temperatures $\taum, \tauv$ and gate $\gate$. The gated mixture is also flexible to implement, since the two terms are normalized and scaled independently.

\section{Experiments}
\label{sec:experiments}

We conduct experiments to quantify the effect of \ours on the long-context capabilities of multi-state linear attention models. We first explain the models, datasets, and baselines used for training and evaluation (\S\ref{sec:setup}). Then, we show the main evaluation results on the core benchmarks (\S\ref{main_results}). Third, we conduct analysis on positional biases and impact of segment length (\S\ref{analysis}). Finally, we conduct ablations on the core design choices in \ours (\S\ref{ablations}).

\begin{table*}[t!]
    \vspace{-5pt}
    \centering    
    \caption{
    \textbf{Performance on Long-Context Retrieval Tasks.} 
    We evaluate mid-trained (\textbf{MidTr}) models with different aggregation methods on the NIAH and QA tasks in RULER at 16K context length, \textbf{--} is the base model without mid-training. Highest scores in each column are in \textbf{bold} and second-highest are \underline{underlined}.
    } 
    \resizebox{\textwidth}{!}{
        \begin{tabular}{cc | c c c c c c | c c | c}
\toprule

& MidTr
& S-NIAH-1
& S-NIAH-2
& S-NIAH-3
& MK-NIAH-1
& MQ-NIAH
& MV-NIAH
& HotpotQA
& SQuADQA
& Avg \\
            
\midrule

\multirow{5}{*}{\shortstack{Gated\\DeltaNet\\1.3B}}
& --
& 18.6	
& 25.2
& 4.2		
& 16.8	
& 5.1	
& 15.8	
& 10.0	
& 8.3	
& 13.0	 \\

& Vanilla 
& 89.0	
& 99.2	
& 10.8	
& 23.4	
& 80.8	
& 70.2	
& 15.0	
& 8.3	
& 49.6 \\

& Position 
&  97.2	
&  96.8	
&  16.0	
&  42.2	
&  82.5	
&  78.4	
&  \underline{16.0}	
&  7.3	
&  54.6\\

& Meanpool
& \textbf{99.6}	
& \underline{99.4}	
& \underline{21.2}	
& \textbf{48.4}	
& \underline{86.6}	
& \underline{85.0}	
& 12.0	
& \underline{10.7}
& \underline{57.9}\\

& \cellcolor{gray!20}\ours
& \cellcolor{gray!20}\underline{99.2}	
& \cellcolor{gray!20}\textbf{100.0}	
& \cellcolor{gray!20}\textbf{47.0}	
& \cellcolor{gray!20}\underline{45.0}	
& \cellcolor{gray!20}\textbf{91.0}	
& \cellcolor{gray!20}\textbf{88.8}	
& \cellcolor{gray!20}\textbf{17.0}	
& \cellcolor{gray!20}\textbf{13.7}	
& \cellcolor{gray!20}\textbf{62.7}\\
            
\midrule
\multirow{5}{*}{\shortstack{Mamba-2\\780M}}
& --
& 0.0	
& 0.0	
& 0.0	
& 0.0
& 0.0
& 0.0	 	
& 2.0
& 0.0
& 0.3 \\

& Vanilla 
& \textbf{100}	
& \textbf{100}	
& 8.4	
& 24.6	
& 92.2	
& 86.4	
& 16	
& 4.7
& 54.0 \\

& Position 
& \textbf{100.0}	
& \textbf{100.0}	
& 54.2	
& \underline{91.0}	
& \underline{94.5}	
& 89.2	
& \underline{20.0}	
& 8.0	
& 69.6 \\

& Meanpool
& \textbf{100.0}	
& \textbf{100.0}	
& \underline{66.2}	
& 89.4	
& 93.7	
& \underline{91.8}	
& \underline{20.0}	
& \underline{8.1}	
& \underline{71.2} \\

& \cellcolor{gray!20}\ours
& \cellcolor{gray!20}\underline{99.8}	
& \cellcolor{gray!20}\underline{99.8}	
& \cellcolor{gray!20}\textbf{96.0}	
& \cellcolor{gray!20}\textbf{94.8}	
& \cellcolor{gray!20}\textbf{95.9}	
& \cellcolor{gray!20}\textbf{92.2}	
& \cellcolor{gray!20}\textbf{22.0}	
& \cellcolor{gray!20}\textbf{9.4}		
& \cellcolor{gray!20}\textbf{76.2} \\

\bottomrule
\end{tabular}
}
\label{tab:ruler}
\end{table*}

\subsection{Setup}
\label{sec:setup}

\paragraph{Models.}
We use the Gated DeltaNet~\citep{yang2025gated} and Mamba-2~\citep{dao2024transformers} architectures for the following experiments. Gated DeltaNet is a linear attention model that updates the memory state following the recurrence in \cref{eq:gdn}. Mamba-2 is a state-space model that uses recurrence of structured semiseparable matrices. We show that \ours improves upon the multi-state variants of both architectures with distinct features to demonstrate its robustness and effectiveness.
A more detailed pseudocode for each model can be found in \cref{app:pseudocode}.

\paragraph{Training.} 
We first pre-train the Gated DeltaNet-1.3B~\citep{yang2025gated} model on 50B tokens in FineWeb-edu~\citep{penedo2024fineweb} following the official training configurations. For Mamba-2-780M, we use the officially released checkpoint\footnote{huggingface state-spaces/mamba2-780m} pre-trained on 300B tokens from Pile~\citep{gao2020pile}. These pre-trained models are still single-state: they require mid-training (or continued pre-training) to function as multi-state variants, \ie, training the pre-trained model with the same objective function as pre-training on the mid-training dataset. 

In order to convert single-state models to multi-state variants, we introduce separation (\cref{eq:separation}) and aggregation (\cref{eq:aggregation}) mechanisms during mid-training. We fix the separation mechanism to uniform segmentation, where each segment has the same length of 256 tokens. At training context length of 16K tokens, the model learns to aggregate as many as 64 (16K / 256) segment-level outputs. We specify the full list of training hyperparameters and model configurations in \cref{app:model-config}.

We mid-train the pre-trained models in two stages. Each stage is defined by the type of training data that was used. Stage 1 uses 2B tokens from Long-Data-Collections (LDC)~\citep{longdatacollections}, with 10\% of the training samples (0.2B tokens) augmented with RULER~\citep{hsieh2024ruler} needle-in-a-haystack (NIAH)-like tasks. The augmented data contains artificial keys and values scattered throughout regular LDC sequences. The task query and ground truth answer are attached at the end of the context. This dataset is designed to train basic language modeling capability as a multi-state variant by activating its aggregation mechanism.   

In stage 2, we train on a smaller dataset of 60M tokens. It consists of 50\% regular LDC sequences, and 50\% synthetically generated samples from RULER HotpotQA and SQuADQA. This dataset is designed to further strengthen the model's aggregation mechanism under reasoning contexts. After stage 1 and stage 2, the pre-trained single-state models now function as multi-state models with long-context capabilities. 
The detailed mid-training hyperparameters can be found in \cref{app:train-config}.

\paragraph{Benchmarks.}
We evaluate the models on various long-context benchmarks. First, we evaluate on the NIAH and QA tasks in RULER~\citep{hsieh2024ruler} at 16K context length. The single needle variants (S-NIAH-1/2/3) evaluate the model's ability to retrieve a single key-value pair in different types of natural and synthetic text haystacks. The multi-needle variants (MK-NIAH-1/MQ-NIAH/MV-NIAH) evaluate the model's ability to retrieve from multiple key-value pairs in different formats. The HotpotQA and SQuADQA tasks in RULER evaluate multi-hop and single-hop reasoning skills. 

We also evaluate on the following subtasks in LongBench~\citep{bai2024longbench}: NarrativeQA (NQ), Qasper (QR), MultiFieldQA-en (MF), HotpotQA (HP), 2WikiMultihopQA (2W), GovReport (GR), QMSum (QM), MultiNews (MN), SAMSum (SS), TREC (TC), TriviaQA (TQ). We filter for samples under 16K context length. These tasks largely evaluate single-document QA, multi-document QA, summarization, and few-shot learning capabilities. 

Finally, we evaluate on the TriviaQA retrieval-augmented-generation task in HELMET~\citep{yen2025helmet} at 16K context length. We use this benchmark to analyze how the position of the ground truth in the sequence impacts the model's performance.
\cref{app:evaluation} provides a detailed explanation of the evaluation protocol used in our experiments.

\paragraph{Baselines.}
We compare \ours with three other baselines: \textbf{Vanilla}, \textbf{Position}, and \textbf{Meanpool}. \textbf{Vanilla} is the unmodified single-state backbone, with neither segmentation nor aggregation mechanisms. The other two are alternative aggregation methods for multi-state variants. \textbf{Position} (\cref{eq:position}) derives aggregation weights from the hidden states alone, scaling segment-level outputs by a learned recency scalar~\citep{guo2026log}. \textbf{Meanpool} (\cref{eq:mc}) derives aggregation weights from first-order alignment between a query and each segment's keys \citep{behrouz2026memory}.

\subsection{Main Results}
\label{main_results}

We report results on RULER~\citep{hsieh2024ruler} NIAH and QA tasks.  \cref{tab:ruler} shows that \ours outperforms baselines in most tasks for both GDN and Mamba-2 architectures. \ours applied to GDN leads to an average score of 62.7, while Meanpool has an average score of 57.9, followed by Position's 54.6. In the case of Mamba-2, \ours shows a 7.0\% relative improvement over Meanpool (76.2 vs. 71.2), and a 9.5\% relative improvement over Position (vs. 69.6). This suggests that mid-training with \ours improves long-context retrieval.

We also report results on LongBench~\citep{bai2024longbench} tasks filtered to 16K. Overall, \ours demonstrates strongest average performance, followed by Meanpool, Position, and finally Vanilla. \cref{tab:longbench} shows that \ours leads to robust gains particularly on single-doc and multi-doc QA. For GDN, \ours leads to a 13.5\% improvement in average score compared to Meanpool (21.9 vs 19.3). For Mamba-2, \ours leads to a 12.0\% (20.6 vs 18.4) improvement compared to Meanpool. This suggests that \ours strengthens general long-context capabilities.

\begin{table*}[t]
    \centering
    \small
    \vspace{-10pt}
    \caption{
    \textbf{Performance on Long-Context Tasks in LongBench.}
    We evaluate on eleven LongBench tasks, filtered for samples with at most 16K tokens. 
    }
    \vspace{-5pt}
    \resizebox{\textwidth}{!}{
        \begin{tabular}{c c | c c c | c c | c c c | c c c | c}
            \toprule
             &
             & \multicolumn{3}{c|}{Single-doc QA} 
             & \multicolumn{2}{c|}{Multi-doc QA} 
             & \multicolumn{3}{c|}{Summarization}
             & \multicolumn{3}{c|}{Few-shot Learning}
             & \\
& MidTr  
& NQ
& QR
& MF
& HP
& 2W
& GR
& QM
& MN
& SS
& TC
& TQ
& Avg \\
\midrule
\multirow{5}{*}{\shortstack{Gated\\DeltaNet\\1.3B}}
& - 
& 2.4	
& 5.0	
& 13.9	
& 4.5	
& 7.9	
& \underline{9.1}	
& \underline{16.1}	
& 9.2	
& 33.0	
& 22.3	
& 19.5	
& 13.0 \\

& Vanilla  
& 6.0	
& 6.1	
& 12.2	
& 6.8	
& 8.5	
& \textbf{9.9}	
& \textbf{16.5}	
& 9.4	
& \textbf{40.5}
& 42.5	
& 21.4	
& 16.3 \\

& Position  
& \underline{11.5}	
& 9.6	
& 14.9
& 14.3	
& \underline{11.1}	
& 6.1	
& 14.0	
& \underline{12.3}	
& 39.0	
& 48.5	
& \textbf{23.1}	
& 18.6 \\

& Meanpool
& 8.8	
& \underline{15.2}	
& \underline{18.0}	
& \underline{14.6}	
& 9.8	
& 8.7	
& 15.3	
& 10.1	
& 34.5	
& \underline{54.8}	
& \underline{22.4}	
& \underline{19.3}  \\

& \cellcolor{gray!20} \ours
& \cellcolor{gray!20}\textbf{13.9}	
& \cellcolor{gray!20}\textbf{19.4}	
& \cellcolor{gray!20}\textbf{20.3}		
& \cellcolor{gray!20}\textbf{16.3}	
& \cellcolor{gray!20}\textbf{14.4}		
& \cellcolor{gray!20}7.8	
& \cellcolor{gray!20}15.9	
& \cellcolor{gray!20}\textbf{12.8}	
& \cellcolor{gray!20}\underline{40.0}	
& \cellcolor{gray!20}\textbf{60.4}	
& \cellcolor{gray!20}20.1	
& \cellcolor{gray!20}\textbf{21.9}\\

\midrule
\multirow{5}{*}{\shortstack{Mamba-2\\780M}}
& - 
& 0.9	
& 1.6	
& 3.6	
& 0.9	
& 2.2	
& 3.2
& 8.7	
& 12.6	
& 8.0	
& 8.1	
& 1.4	
& 4.7	 \\

& Vanilla  
& 1.0	
& 3.2	
& 7.6	
& 4.2	
& 5.2	
& 10.2	
& 12.3	
& 14.7	
& 36.5	
& 54.5	
& 5.7	
& 14.1 \\

& Position  
& 1.2	
& \underline{5.1}	
& 10.5	
& 5.7	
& \underline{6.8}	
& 15.5	
& 15.0	
& 17.4	
& 42.0	
& \underline{65.1}	
& 12.5	
& 17.9 \\

& Meanpool
& \underline{1.7}	
& 4.2	
& \underline{11.5}	
& \underline{6.0}	
& 5.6	
& \underline{16.0}	
& \textbf{17.7}
& \underline{18.5}	
& \underline{44.5}	
& 61.1	
& \textbf{15.1} 
& \underline{18.4} \\

& \cellcolor{gray!20} \ours
& \cellcolor{gray!20}\textbf{1.9}	
& \cellcolor{gray!20}\textbf{6.7}	
& \cellcolor{gray!20}\textbf{12.6}	
& \cellcolor{gray!20}\textbf{7.8}	
& \cellcolor{gray!20}\textbf{9.1}	
& \cellcolor{gray!20}\textbf{20.4}	
& \cellcolor{gray!20}\underline{16.1}	
& \cellcolor{gray!20}\textbf{19.2}	
& \cellcolor{gray!20}\textbf{46.0}	
& \cellcolor{gray!20}\textbf{72.7}	
& \cellcolor{gray!20}\underline{14.2}	
& \cellcolor{gray!20}\textbf{20.6} \\

\bottomrule
\end{tabular}
}
\label{tab:longbench}
\end{table*}

\begin{figure}[t]

\centering
\begin{subfigure}[t]{0.49\textwidth}
    \centering
    \includegraphics[width=\textwidth]{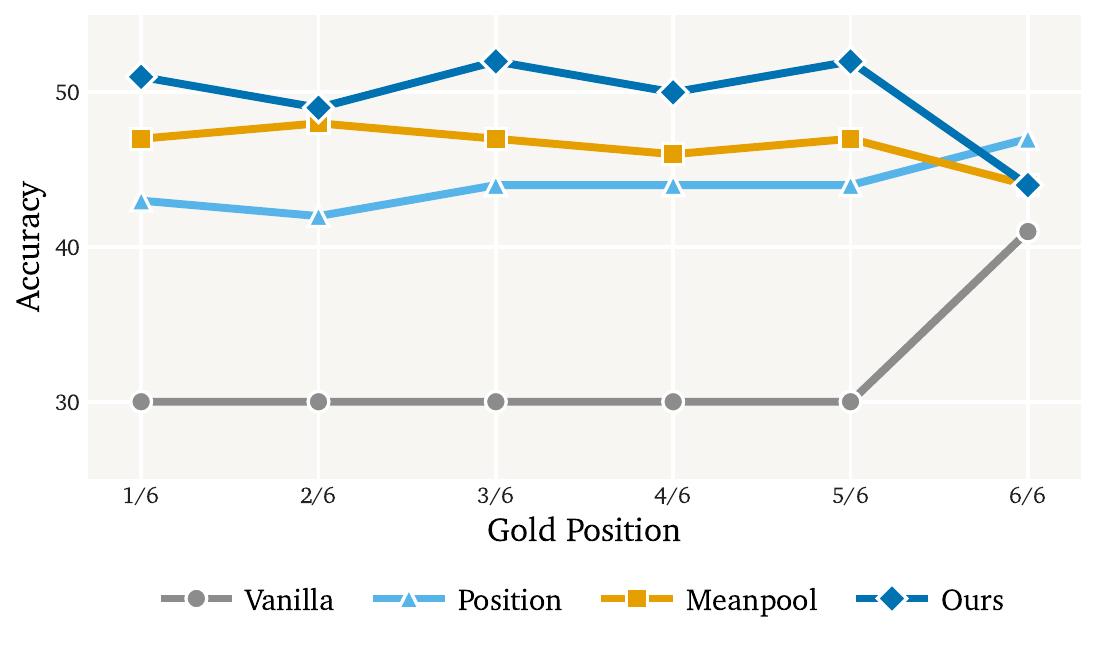}
    \caption{Gated DeltaNet-1.3B}
    \label{fig:position_gdn}
\end{subfigure}
\hfill
\begin{subfigure}[t]{0.49\textwidth}
    \centering
    \includegraphics[width=\textwidth]{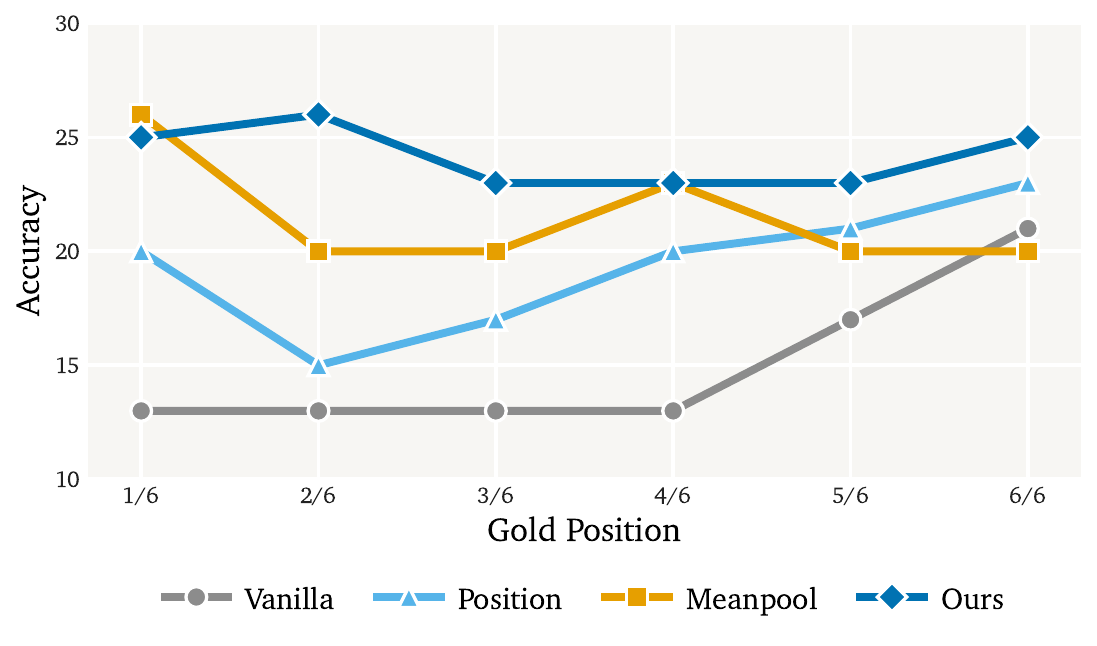}
    \caption{Mamba-2-780M}
    \label{fig:position_mamba}
\end{subfigure}
\caption{\textbf{Position-Controlled Retrieval.} Retrieval accuracy as the gold passage sweeps from the start ($1/6$) to the end ($6/6$) of the context. Vanilla is recency-biased: it recovers the gold only when placed near the end and is otherwise flat, while Position and mean-based aggregation are flatter but weaker. \ours stays high and roughly uniform across positions.}
\label{fig:position}
\end{figure}

\begin{figure}[t]
\centering
\begin{subfigure}[t]{0.49\textwidth}
    \centering
    \includegraphics[width=\textwidth]{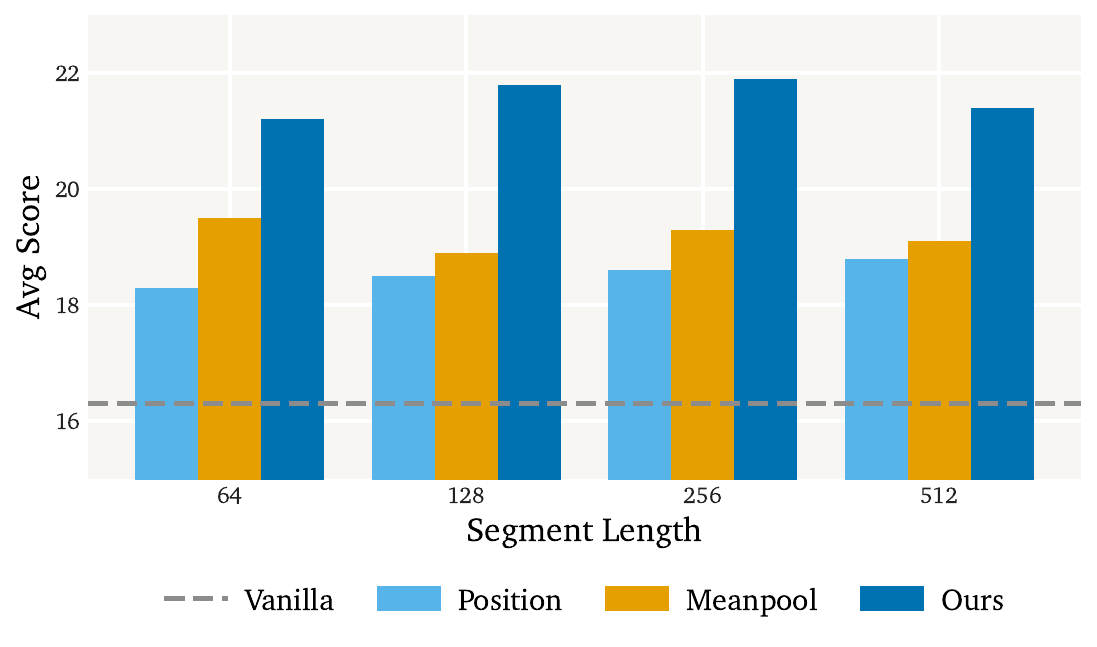}
    \caption{Gated DeltaNet-1.3B}
    \label{fig:segment_gdn}
\end{subfigure}
\hfill
\begin{subfigure}[t]{0.49\textwidth}
    \centering
    \includegraphics[width=\textwidth]{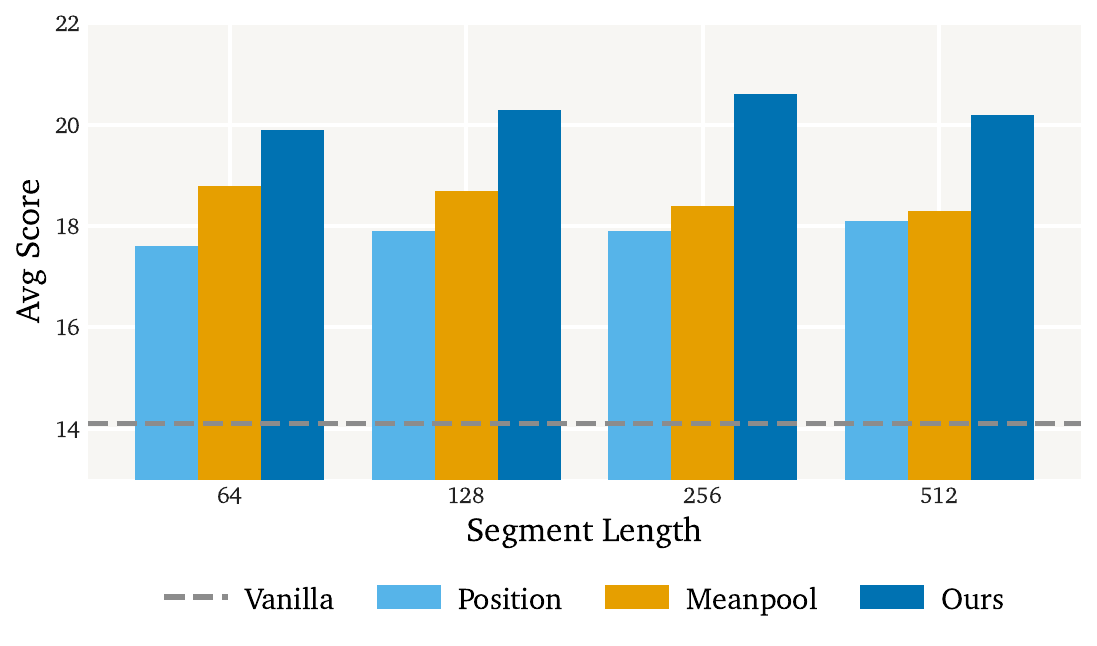}
    \caption{Mamba-2-780M}
    \label{fig:segment_mamba}
\end{subfigure}
\caption{\textbf{\ours Outperforms Baselines Across Segment Length.} Average score as the segment size that defines each state varies over $\{64,128,256,512\}$. \ours leads at every segment length, and its margin over mean-based aggregation persists as segments grow. Vanilla (dashed) is segment-independent.}
\label{fig:segment}
\end{figure}

\subsection{Analysis} 
\label{analysis}

\paragraph{Positional Bias Analysis.}
We evaluate models on the TriviaQA task in HELMET~\citep{yen2025helmet}, at 16K context length. This is a retrieval-augmented generation task, where the context consists of multiple passages and only one of them, which we refer to as the gold passage, is related to the final query of the prompt. The location of the gold passage is shuffled randomly within the context for each sample. Therefore, we can track the accuracy of the model's responses depending on the location of the gold passage, and detect any positional biases that might exist. \cref{fig:position} shows that the Vanilla baseline demonstrates a strong recency bias: it struggles unless the gold passage is located towards the end of the context. The Position baseline also demonstrates a weak recency bias, where the accuracy gradually increases as the gold passage moves closer to the end. Both the Meanpool baseline and \ours perform more evenly across gold passage positions, while \ours is better overall. This suggests that \ours enhances the model's ability to extract information from important segments in the context by improving aggregation of segment-level readouts.

\paragraph{Segment Length Analysis.} 
We examine whether \ours is robust to different segment lengths. Segment length, or the number of tokens per segment, determines the number of segments per context given that the entire length of the sequence is fixed ($L = T / m$). The shorter the segment length, the more states per sequence, and therefore the higher the memory consumption of the model. We experiment with $m \in \{64, 128, 256, 512\}$ for each aggregation method. At $m = 64$, the model consumes approximately 8 times more memory compared to $m = 512$. \cref{fig:segment} shows that for both GDN and Mamba-2, \ours outperforms baselines on LongBench~\citep{bai2024longbench} tasks across $m$. In particular, \ours at $m = 256$ (default) outperforms Meanpool and Position baselines at $m=64$, which consume 4 times more memory. This suggests that \ours is more memory efficient compared to its baselines. 

\paragraph{Memory Cost and Parameter Overhead.}
In ~\cref{app:additional-analysis}, we compare the memory cost of different multi-state linear attention strategies with \ours. We observe that the extra cost of our additional second moment features is limited to at most 12.6\% for Gated DeltaNet and 4.2\% for Mamba-2 at $d_r=128$, while the memory cost decreases quadratically with $d_r$. We also compare the number of parameters that \ours adds to each backbone compared to baselines. We find that the \ours's parameter overhead is on par with the Meanpool baseline at 4.6\% for GDN and 2.2\% for Mamba-2. 

\paragraph{Qualitative Examples.}
In Appendix~\S\ref{app:additional-analysis}, we provide examples of aggregation weight ($\lambda_{t, i}$) assignment on four HELMET~\citep{yen2025helmet} retrieval-augmented generation samples using \ours and baseline methods. As observed in the quantitative evaluations, \ours assigns a higher weight to the gold segment, and this supports that the variance of the alignment is critical for aggregation.

\subsection{Ablations}
\label{ablations}

\begin{wrapfigure}{r}{0.45\textwidth}
    \centering
    \vspace{-10pt}
    \includegraphics[width=0.44\textwidth]{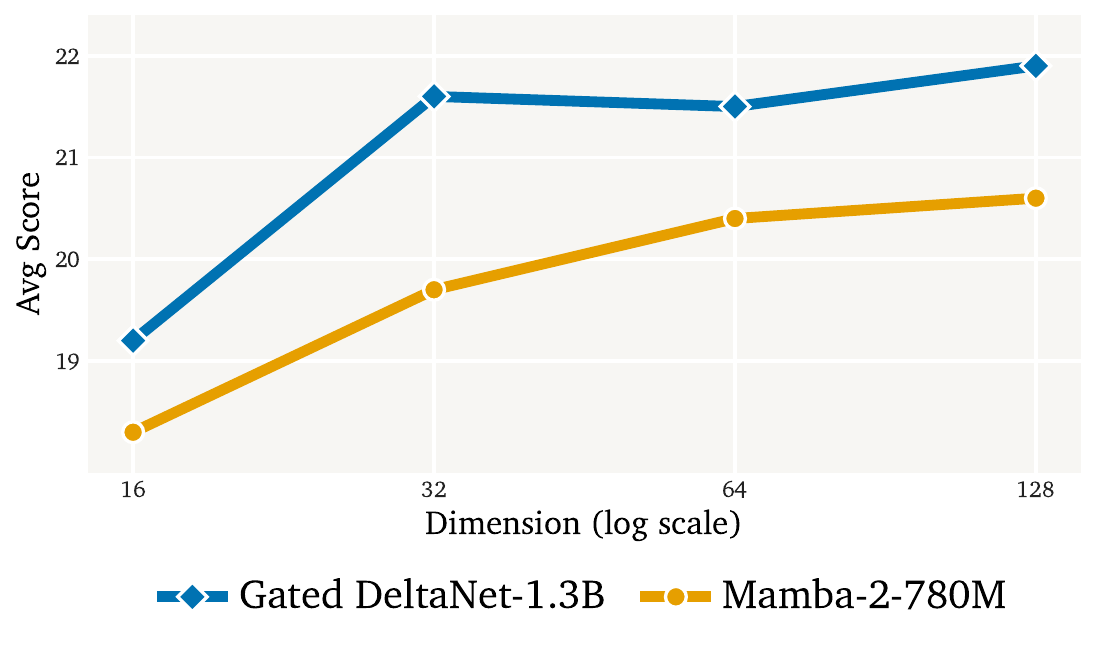}
    \vspace{-10pt}
    \caption{\small \textbf{Dimension of Moments.} Average LongBench scores as $\rdim$ varies over $\{16,32,64,128\}$ (log axis). Performance saturates around $\rdim\,{=}\,32$ for GDN and $\rdim\,{=}\,64$ for Mamba-2.
    }
    \label{fig:dim}
\end{wrapfigure}
\paragraph{Ablations on Dimension of Moments of Keys.}
We report on the impact of $d_r$, the dimension of the first and second moments of keys (\S\ref{sec:weight_derivation}), on the performance of \ours. We vary $\rdim \in \{16, 32, 64, 128\}$ during training and evaluate on LongBench~\citep{bai2024longbench}. \cref{fig:dim} shows that the average score on LongBench tasks gradually increases as $\rdim$ increases. This suggests that the width of the moments of keys is critical in deriving informative aggregation weights. However, there is a trade-off in deciding $\rdim$. The moments of keys can be more expressive when $\rdim$ is larger. However, \ours consumes less memory compared to the Meanpool and Position baselines as $\rdim$ gets smaller. More details on the memory consumption can be found in \cref{tab:memory} of \cref{app:additional-analysis}.

\begin{wraptable}{r}{0.45\textwidth}
\centering
\vspace{-\intextsep}
\setlength{\tabcolsep}{4pt}
\scriptsize
\resizebox{\linewidth}{!}{%
\begin{tabular}{lccccc}
\toprule
 & Single & Multi & Summ & Few-shot & Overall \\
\midrule
GDN + \ours      & \textbf{17.9} & \textbf{15.4} & \textbf{12.2} & \textbf{40.2} & \textbf{21.9} \\
\quad + write-forget & 14.9 & 9.2 & 11.6 & 35.3 & 18.5 \\
\quad + shared-KQ    & 10.7 & 9.2 & 10.9 & 29.8 & 15.7 \\
\midrule
Mamba + \ours    & \textbf{7.1} & \textbf{8.5} & \textbf{18.6} & \textbf{44.3} & \textbf{20.6} \\
\quad + write-forget & 6.7 & 6.9 & 16.6 & 43.8	 & 19.6 \\
\quad + shared-KQ    & 5.1 & 6.1 & 15.1 & 38.3 & 17.1 \\
\bottomrule
\end{tabular}%
}
\caption{\small \textbf{Parameter Sharing.} We train two \ours variants: sharing the keys and queries between the memory state and moments of keys (+shared KQ) and using write-forget gates on moments of keys (+write-forget). We evaluate on LongBench and report per-category average scores. The categories are identical to \cref{tab:longbench}.}
\vspace{-1em}
\label{tab:paramshare}
\end{wraptable}

\paragraph{Ablations on Parameter Sharing.}
For each token, we project new keys and queries ($k'_t, q'_t$) to construct the first and second moments, separate from the original keys and queries for the memory state ($k_t, q_t$). We isolate the impact of decoupling the keys and queries by experimenting with a variant of \ours  where  $k'_t = k_t$ and $q'_t = q_t$ (shared-KQ rows in \cref{tab:paramshare}). 
We further experiment with a variant of \ours where we control the decay and update strength of the first and second moments of keys per token, similar to the decay and write-strength features of the memory state explained in \cref{eq:gdn} (write-forget rows in \cref{tab:paramshare}). We find that both ablations substantially degrade the performance of \ours on LongBench~\citep{bai2024longbench}. This suggests that using a dedicated set of keys and queries from moments of keys is particularly important in deriving informative aggregation weights.

\section{Conclusion}

We investigate the aggregation problem in multi-state linear attention, where segment-level outputs are combined via a weighted sum to produce the final output representation of each token. We propose \ours, an aggregation method that assigns weights to each segment based on the mean and variance of the alignment between a query and the keys of each segment. In order to compute the mean and variance of the alignment, \ours keeps compact first and second moments of the segment's keys. We find that \ours is superior to existing aggregation methods in terms of performance and memory efficiency. \ours presents a generalizable and effective routing mechanism for the aggregation problem in multi-state linear attention models. 

\ificlrfinal
\subsection*{Acknowledgments}
All experiments, data collection, and processing activities were conducted at Princeton University, and no Google DeepMind computing resources were used for this work.
Sanghyuk Chun is supported by the Francis Robbins Upton Fellowship at Princeton University.
Xindi Wu is supported by the Apple Fellowship.
\fi

\subsection*{AI use statement}

In this work, we used generative AI tools to polish the writing, generate synthetic training datasets (\S\ref{sec:setup}), and write triton kernels for method implementation. We also used AI tools to plot figures and create a name for our method. We have not used generative AI tools to formulate and develop the conceptual framework of the method, or to identify relevant literature. We have reviewed all AI-assisted work. We take responsibility for the final content of this work,
including text, claims or artifacts produced with the aid of generative AI.

\subsection*{Ethics statement}

Our work presents a method that improves general long-context capabilities of multi-state linear attention models. Although we do not see a clear ethical concern, we advise readers to be cautious with the datasets and tools they may use to extend our work beyond the scope of this paper.  

\subsection*{Reproducibility statement}

We use publicly available model architectures and datasets for training and evaluation. We discuss the method details in \cref{sec:method}, and specifics of the training procedure in \cref{sec:setup}. We also provide a full summary of model configurations and training hyperparameters in Appendix~\cref{app:config}. The evaluation details can be found in \cref{app:evaluation}.

\bibliography{iclr2027_conference}

@inproceedings{katharopoulos2020transformers,
  title={Transformers are rnns: Fast autoregressive transformers with linear attention},
  author={Katharopoulos, Angelos and Vyas, Apoorv and Pappas, Nikolaos and Fleuret, Fran{\c{c}}ois},
  booktitle={International conference on machine learning},
  pages={5156--5165},
  year={2020},
  organization={PMLR}
}

@article{wang2026dynamic,
  title={Dynamic Linear Attention},
  author={Wang, Xin and Shen, Hui and Zheng, Boyuan and Liu, Xueshen and Cho, Minkyoung and Wan, Zhongwei and Zhao, Zesen and Mao, Zhuoqing and Yan, Shen and Zhang, Mi},
  journal={arXiv preprint arXiv:2606.10650},
  year={2026}
}

@inproceedings{guo2026log,
  title={Log-linear attention},
  author={Guo, Guo and Yang, Songlin and Goel, Tarushii and Xing, Eric P and Dao, Tri and Kim, Yoon},
  booktitle={International Conference on Learning Representations},
  volume={2026},
  pages={91296--91317},
  year={2026}
}

@article{behrouz2026memory,
  title={Memory Caching: RNNs with Growing Memory},
  author={Behrouz, Ali and Li, Zeman and Deng, Yuan and Zhong, Peilin and Razaviyayn, Meisam and Mirrokni, Vahab},
  journal={arXiv preprint arXiv:2602.24281},
  year={2026}
}

@article{dao2024transformers,
  title={Transformers are SSMs: Generalized models and efficient algorithms through structured state space duality},
  author={Dao, Tri and Gu, Albert},
  journal={arXiv preprint arXiv:2405.21060},
  year={2024}
}

@article{yang2024parallelizing,
  title={Parallelizing linear transformers with the delta rule over sequence length},
  author={Yang, Songlin and Wang, Bailin and Zhang, Yu and Shen, Yikang and Kim, Yoon},
  journal={Advances in neural information processing systems},
  volume={37},
  pages={115491--115522},
  year={2024}
}

@inproceedings{yang2025gated,
  title={Gated delta networks: Improving mamba2 with delta rule},
  author={Yang, Songlin and Kautz, Jan and Hatamizadeh, Ali},
  booktitle={International Conference on Learning Representations},
  volume={2025},
  pages={29687--29707},
  year={2025}
}

@article{penedo2024fineweb,
  title={The fineweb datasets: Decanting the web for the finest text data at scale},
  author={Penedo, Guilherme and Kydl{\'\i}{\v{c}}ek, Hynek and Lozhkov, Anton and Mitchell, Margaret and Raffel, Colin and Von Werra, Leandro and Wolf, Thomas and others},
  journal={Advances in Neural Information Processing Systems},
  volume={37},
  pages={30811--30849},
  year={2024}
}

@misc{longdatacollections,
  title = {Long data collections database},
  author={TogetherAI},
  howpublished = {\url{https://huggingface.co/datasets/togethercomputer/Long-Data-Collections}},
  year={2024}
}

@inproceedings{schlag2021linear,
  title={Linear transformers are secretly fast weight programmers},
  author={Schlag, Imanol and Irie, Kazuki and Schmidhuber, J{\"u}rgen},
  booktitle={International conference on machine learning},
  pages={9355--9366},
  year={2021},
  organization={PMLR}
}

@article{yang2023gated,
  title={Gated linear attention transformers with hardware-efficient training},
  author={Yang, Songlin and Wang, Bailin and Shen, Yikang and Panda, Rameswar and Kim, Yoon},
  journal={arXiv preprint arXiv:2312.06635},
  year={2023}
}

@article{gu2023mamba,
  title={Mamba: Linear-time sequence modeling with selective state spaces},
  author={Gu, Albert and Dao, Tri},
  journal={arXiv preprint arXiv:2312.00752},
  year={2023}
}

@inproceedings{peng2023rwkv,
  title={Rwkv: Reinventing rnns for the transformer era},
  author={Peng, Bo and Alcaide, Eric and Anthony, Quentin and Albalak, Alon and Arcadinho, Samuel and Biderman, Stella and Cao, Huanqi and Cheng, Xin and Chung, Michael and Derczynski, Leon and others},
  booktitle={Findings of the association for computational linguistics: EMNLP 2023},
  pages={14048--14077},
  year={2023}
}

@article{sun2023retentive,
  title={Retentive network: A successor to transformer for large language models},
  author={Sun, Yutao and Dong, Li and Huang, Shaohan and Ma, Shuming and Xia, Yuqing and Xue, Jilong and Wang, Jianyong and Wei, Furu},
  journal={arXiv preprint arXiv:2307.08621},
  year={2023}
}

@inproceedings{dai2019transformer,
  title={Transformer-xl: Attentive language models beyond a fixed-length context},
  author={Dai, Zihang and Yang, Zhilin and Yang, Yiming and Carbonell, Jaime G and Le, Quoc and Salakhutdinov, Ruslan},
  booktitle={Proceedings of the 57th annual meeting of the association for computational linguistics},
  pages={2978--2988},
  year={2019}
}

@article{rae2019compressive,
  title={Compressive transformers for long-range sequence modelling},
  author={Rae, Jack W and Potapenko, Anna and Jayakumar, Siddhant M and Lillicrap, Timothy P},
  journal={arXiv preprint arXiv:1911.05507},
  year={2019}
}

@article{bulatov2022recurrent,
  title={Recurrent memory transformer},
  author={Bulatov, Aydar and Kuratov, Yury and Burtsev, Mikhail},
  journal={Advances in neural information processing systems},
  volume={35},
  pages={11079--11091},
  year={2022}
}

@inproceedings{du2026mom,
  title={Mom: Linear sequence modeling with mixture-of-memories},
  author={Du, Jusen and Sun, Weigao and Lan, Disen and Hu, Jiaxi and Tao, Zhang and Cheng, Yu},
  booktitle={International Conference on Learning Representations},
  volume={2026},
  pages={106613--106631},
  year={2026}
}

@article{pan2025scaling,
  title={Scaling linear attention with sparse state expansion},
  author={Pan, Yuqi and An, Yongqi and Li, Zheng and Chou, Yuhong and Zhu, Ruijie and Wang, Xiaohui and Wang, Mingxuan and Wang, Jinqiao and Li, Guoqi},
  journal={arXiv preprint arXiv:2507.16577},
  year={2025}
}

@article{shazeer2017outrageously,
  title={Outrageously large neural networks: The sparsely-gated mixture-of-experts layer},
  author={Shazeer, Noam and Mirhoseini, Azalia and Maziarz, Krzysztof and Davis, Andy and Le, Quoc and Hinton, Geoffrey and Dean, Jeff},
  journal={arXiv preprint arXiv:1701.06538},
  year={2017}
}

@article{fedus2021switch,
  title={Switch transformers: Scaling to trillion parameter models with simple and efficient sparsity. arXiv 2021},
  author={Fedus, William and Zoph, Barret and Shazeer, Noam},
  journal={arXiv preprint arXiv:2101.03961},
  year={2021}
}

@article{graves2014neural,
  title={Neural turing machines},
  author={Graves, Alex and Wayne, Greg and Danihelka, Ivo},
  journal={arXiv preprint arXiv:1410.5401},
  year={2014}
}

@article{sukhbaatar2015end,
  title={End-to-end memory networks},
  author={Sukhbaatar, Sainbayar and Weston, Jason and Fergus, Rob and others},
  journal={Advances in neural information processing systems},
  volume={28},
  year={2015}
}

@article{wu2022memorizing,
  title={Memorizing transformers},
  author={Wu, Yuhuai and Rabe, Markus N and Hutchins, DeLesley and Szegedy, Christian},
  journal={arXiv preprint arXiv:2203.08913},
  year={2022}
}

@inproceedings{arora2024zoology,
  title={Zoology: Measuring and improving recall in efficient language models},
  author={Arora, Simran and Eyuboglu, Sabri and Timalsina, Aman and Johnson, Isys and Poli, Michael and Zou, James Y and Rudra, Atri and R{\'e}, Christopher},
  booktitle={International conference on learning representations},
  volume={2024},
  pages={15664--15730},
  year={2024}
}

@article{arora2024simple,
  title={Simple linear attention language models balance the recall-throughput tradeoff},
  author={Arora, Simran and Eyuboglu, Sabri and Zhang, Michael and Timalsina, Aman and Alberti, Silas and Zinsley, Dylan and Zou, James and Rudra, Atri and R{\'e}, Christopher},
  journal={arXiv preprint arXiv:2402.18668},
  year={2024}
}

@article{jelassi2024repeat,
  title={Repeat after me: Transformers are better than state space models at copying},
  author={Jelassi, Samy and Brandfonbrener, David and Kakade, Sham M and Malach, Eran},
  journal={arXiv preprint arXiv:2402.01032},
  year={2024}
}

@inproceedings{nichani2025understanding,
  title={Understanding factual recall in transformers via associative memories},
  author={Nichani, Eshaan and Lee, Jason and Bietti, Alberto},
  booktitle={International Conference on Learning Representations},
  volume={2025},
  pages={4166--4207},
  year={2025}
}

@article{hsieh2024ruler,
  title={RULER: What's the real context size of your long-context language models?},
  author={Hsieh, Cheng-Ping and Sun, Simeng and Kriman, Samuel and Acharya, Shantanu and Rekesh, Dima and Jia, Fei and Zhang, Yang and Ginsburg, Boris},
  journal={arXiv preprint arXiv:2404.06654},
  year={2024}
}

@inproceedings{bai2024longbench,
  title={Longbench: A bilingual, multitask benchmark for long context understanding},
  author={Bai, Yushi and Lv, Xin and Zhang, Jiajie and Lyu, Hongchang and Tang, Jiankai and Huang, Zhidian and Du, Zhengxiao and Liu, Xiao and Zeng, Aohan and Hou, Lei and others},
  booktitle={Proceedings of the 62nd annual meeting of the association for computational linguistics (volume 1: Long papers)},
  pages={3119--3137},
  year={2024}
}

@inproceedings{yen2025helmet,
  title={HELMET: How to evaluate long-context models effectively and thoroughly},
  author={Yen, Howard and Gao, Tianyu and Hou, Minmin and Ding, Ke and Fleischer, Daniel and Izsak, Peter and Wasserblat, Moshe and Chen, Danqi},
  booktitle={International Conference on Learning Representations},
  volume={2025},
  pages={98914--98965},
  year={2025}
}

@article{vaswani2017attention,
  title={Attention is all you need},
  author={Vaswani, Ashish and Shazeer, Noam and Parmar, Niki and Uszkoreit, Jakob and Jones, Llion and Gomez, Aidan N and Kaiser, {\L}ukasz and Polosukhin, Illia},
  journal={Advances in neural information processing systems},
  volume={30},
  year={2017}
}

@article{gao2020pile,
  title={The pile: An 800gb dataset of diverse text for language modeling},
  author={Gao, Leo and Biderman, Stella and Black, Sid and Golding, Laurence and Hoppe, Travis and Foster, Charles and Phang, Jason and He, Horace and Thite, Anish and Nabeshima, Noa and others},
  journal={arXiv preprint arXiv:2101.00027},
  year={2020}
}

@inproceedings{keles2023computational,
  title={On the computational complexity of self-attention},
  author={Keles, Feyza Duman and Wijewardena, Pruthuvi Mahesakya and Hegde, Chinmay},
  booktitle={International conference on algorithmic learning theory},
  pages={597--619},
  year={2023},
  organization={PMLR}
}

@article{kuratov2024babilong,
  title={Babilong: Testing the limits of llms with long context reasoning-in-a-haystack},
  author={Kuratov, Yuri and Bulatov, Aydar and Anokhin, Petr and Rodkin, Ivan and Sorokin, Dmitry and Sorokin, Artyom and Burtsev, Mikhail},
  journal={Advances in Neural Information Processing Systems},
  volume={37},
  pages={106519--106554},
  year={2024}
}

@article{zhang2024gated,
  title={Gated slot attention for efficient linear-time sequence modeling},
  author={Zhang, Yu and Yang, Songlin and Zhu, Ruijie and Zhang, Yue and Cui, Leyang and Wang, Yiqiao and Wang, Bolun and Shi, Freda and Wang, Bailin and Bi, Wei and others},
  journal={Advances in Neural Information Processing Systems},
  volume={37},
  pages={116870--116898},
  year={2024}
}

@misc{eval-harness,
  author       = {Gao, Leo and Tow, Jonathan and Abbasi, Baber and Biderman, Stella and Black, Sid and DiPofi, Anthony and Foster, Charles and Golding, Laurence and Hsu, Jeffrey and Le Noac'h, Alain and Li, Haonan and McDonell, Kyle and Muennighoff, Niklas and Ociepa, Chris and Phang, Jason and Reynolds, Laria and Schoelkopf, Hailey and Skowron, Aviya and Sutawika, Lintang and Tang, Eric and Thite, Anish and Wang, Ben and Wang, Kevin and Zou, Andy},
  title        = {The Language Model Evaluation Harness},
  month        = 07,
  year         = 2024,
  publisher    = {Zenodo},
  version      = {v0.4.3},
  doi          = {10.5281/zenodo.12608602}
}

@article{hwang2026reinforced,
  title={Reinforced Fast Weights with Next-Sequence Prediction},
  author={Hwang, Hee Seung and Wu, Xindi and Chun, Sanghyuk and Russakovsky, Olga},
  journal={arXiv preprint arXiv:2602.16704},
  year={2026}
}

@article{wu2026addressable,
  title={Addressable Memory for Video World Models},
  author={Wu, Xindi and Elflein, Sven and Lucas, James and Russakovsky, Olga and Leal-Taix{\'e}, Laura and Paschalidou, Despoina and Lorraine, Jonathan and O{\v{s}}ep, Aljo{\v{s}}a},
  journal={arXiv preprint arXiv:2608.07408},
  year={2026}
}
\bibliographystyle{iclr2027_conference}

\clearpage
\makeatletter
\let\addcontentsline\BD@acl
\makeatother

\hypertarget{toc}{}
\begingroup
\makeatletter
\setcounter{tocdepth}{2}
\titlecontents{section}[1.6em]{\addvspace{0.9em}\bfseries\color{linkblue}}%
  {\contentslabel{1.6em}}{\hspace*{-1.6em}}{\hfill\bfseries\color{black}\contentspage}
\titlecontents{subsection}[4em]{\color{linkblue}}%
  {\contentslabel{2.4em}}{\hspace*{-2.4em}}{\titlerule*[0.6pc]{.}\color{black}\contentspage}
\noindent{\fontsize{28}{32}\selectfont\bfseries Appendices}\par\vspace{0.8em}
\@starttoc{toc}
\makeatother
\endgroup
\vspace{1em}

\appendix

\pagestyle{fancy}
\fancyfoot{}
\fancyfoot[L]{\hyperlink{toc}{Back to Table of Contents}}
\fancyfoot[C]{\thepage}
\fancyfoot[R]{\hyperlink{abstract}{Back to the First Page}}

\section{Notation}
\label{app:notation}

\begingroup
\renewcommand{\arraystretch}{1.4}
\begin{xltabular}{\textwidth}{@{}l X@{}}
\caption{Glossary and Notation.}
\label{tab:notation}\\
\toprule
\textbf{Symbol} & \textbf{Description} \\
\midrule
\endfirsthead
\multicolumn{2}{@{}l}{\textit{\small Table~\ref{tab:notation} (continued)}} \\
\toprule
\textbf{Symbol} & \textbf{Description} \\
\midrule
\endhead
\midrule
\multicolumn{2}{r@{}}{\textit{\small continued on next page}} \\
\endfoot
\bottomrule
\endlastfoot
\multicolumn{2}{@{}l}{\textit{\textcolor{gray}{Sequences and softmax attention}}} \\
$\seqlen$ & Sequence / context length (number of tokens) \\
$t$ & Position of the target token \\
$\vx_t$ & Hidden state of token $t$ \\
$\hdim$ & Key/value dimension of one head's memory state (written $\hdim$ for both, whereas the backbones use $d_k{\times}d_v$, see \cref{tab:memory}) \\
$\mQ,\mK,\mV \in \mathbb{R}^{\seqlen\times \hdim}$ & Query, key, and value matrices \\
$\vq_t,\vk_t,\vv_t$ & Query, key, and value of token $t$ \\
$\mW_q,\ \mW_k,\ \mW_v$ & Backbone projections producing $\vq_t,\vk_t,\vv_t$ from $\vx_t$ \\
$\mM$ & Causal attention mask. In softmax attention it is additive with entries in $\{-\infty,0\}$. In linear attention it is binary, with $\mM_{ij}=1$ if $i\ge j$ and $0$ otherwise \\
$\mO,\ \vo_t$ & Attention output, as a matrix and for token $t$ \\
$\softmax$ & Row-wise softmax normalization \\
$\odot$ & Elementwise (Hadamard) product \\
$\mI$ & Identity matrix \\
\addlinespace[0.4em]
\multicolumn{2}{@{}l}{\textit{\textcolor{gray}{Linear attention}}} \\
$\mS_t \in \mathbb{R}^{\hdim\times \hdim}$ & Recurrent memory state at step $t$ \\
$\decay \in (0,1)$ & State decay gate in Gated DeltaNet (\cref{eq:gdn}) \\
$\beta_t$ & Write strength of the delta term (\cref{eq:gdn}) \\
\addlinespace[0.4em]
\multicolumn{2}{@{}l}{\textit{\textcolor{gray}{Multi-state recurrent memory}}} \\
$\nseg$ & Number of segments (memory states) \\
$\seg$ & The $i$-th context segment \\
$i,\ \ell$ & Segment indices ($1 \le i,\ell \le \nseg$) \\
$j$ & Index of a token within a segment ($j \in \seg$) \\
$\segsize = |\seg|$ & Number of tokens per segment \\
$\segstate \in \mathbb{R}^{\hdim\times \hdim}$ & Memory state of segment $i$. For the plain linear recurrence, $\segstate=\sum_{j\in \seg}\vv_j\vk_j^\top$ \\
$\vo_{t,i}$ & Segment-level output at position $t$: readout of segment $i$'s state, $\vo_{t,i}=\segstate\vq_t$ \\
$\aggw$ & Aggregation weight of segment $i$ at position $t$ \\
$\boldsymbol{\lambda}_t,\ \mW_\lambda$ & Position-based weight vector (entries $\aggw$) and its projection \\
$\mW_u,\ \vu_t$ & Target-token projection and vector used by mean-key routing \\
$\bar{\vk}_i$ & Meanpooled key of segment $i$ \\
\addlinespace[0.4em]
\multicolumn{2}{@{}l}{\textit{\textcolor{gray}{\ours: key moments and aggregation weights}}} \\
$\rdim$ & Moment dimension (hyperparameter) \\
$\rquery,\ \rkey \in \mathbb{R}^{\rdim}$ & New query and key of token $t$, projected from its hidden state \\
$\mW_{\vq'},\ \mW_{\vk'}$ & Projections producing $\rquery$ and $\rkey$ from $\vx_t$ \\
$\firstm \in \mathbb{R}^{\rdim}$ & First moment of segment $i$'s keys (mean of $\rkey$ over $\seg$) \\
$\secondm \in \mathbb{R}^{\rdim\times \rdim}$ & Uncentered second moment of segment $i$'s keys (mean of $\rkey{\rkey}^\top$ over $\seg$) \\
$\rmean$ & Mean key--query alignment for segment $i$, $\rmean=\firstm^\top \rquery$ \\
$\rvar$ & Variance of the key--query alignment over segment $i$, $\rvar={\rquery}^\top\secondm \rquery-(\rmean)^2$ \\
$\gate$ & Learned gate mixing the softmax-normalized mean and variance terms (\cref{eq:weight}), distinct from the decay gate above \\
$\taum,\ \tauv$ & Learnable temperatures for the mean term ($\taum$) and the variance term ($\tauv$) \\
\addlinespace[0.4em]
\multicolumn{2}{@{}l}{\textit{\textcolor{gray}{Connection to softmax attention}}} \\
$s_{t,i,j}$ & Query-key similarity, $s_{t,i,j}={\rquery}^\top \vk'_j$ for $j\in \seg$ \\
$\delta$ & Temperature in the grouped-softmax analysis (\S\ref{sec:connection}) \\
$\gsZ(\delta)$ & Length-normalized softmax evidence of segment $i$ \\
$\gsA$ & Grouped softmax attention: total softmax mass on segment $i$ \\
$\gsmu$ & Mean of $s_{t,i,j}$ over $\seg$ (first cumulant) \\
$\gsvar$ & Variance of $s_{t,i,j}$ over $\seg$ (second cumulant) \\
\addlinespace[0.4em]
\multicolumn{2}{@{}l}{\textit{\textcolor{gray}{Mamba-2 backbone (pseudocode)}}} \\
$P,\ N$ & Head dimension and SSM state size, with $\segstate\in\mathbb{R}^{P\times N}$ \\
$\Delta_t,\ A$ & Step size at position $t$ and per-head scalar decay rate, so the state decays by $\exp(\Delta_t A)$ \\
$\mB_t,\ \mD_t$ & Input matrix (the write key) and output matrix (the read query, written $\mC$ in Mamba-2) \\
$\mW_B,\ \mW_D,\ \mW_\Delta$ & Projections producing $\mB_t$, $\mD_t$, and $\Delta_t$ from $\vx_t$ \\
\end{xltabular}
\endgroup

\section{Related Work}
\label{app:relwork}

\paragraph{Linear Attention and Recurrent Memory Models.}
Linear attention replaces the softmax normalization in standard attention with a feature map, turning attention into a recurrence with a fixed-size memory state~\citep{katharopoulos2020transformers}. This state can be viewed as a fast-weight associative memory written by outer products~\citep{schlag2021linear,hwang2026reinforced}. Recent works improve this recurrence with data-dependent gating~\citep{yang2023gated}, delta-rule updates that overwrite rather than accumulate~\citep{yang2024parallelizing}, and their combination~\citep{yang2025gated}. Selective state-space models~\citep{gu2023mamba,dao2024transformers}, and recurrent memory models such as RWKV~\citep{peng2023rwkv} and RetNet~\citep{sun2023retentive} that are closely related to linear attention share the same fixed-size state form. These models compress the context into a single state, which may be limited in memory capacity due to information loss at longer contexts. Our work builds on these backbones to train more capable multi-state linear attention and recurrent memory models.

\paragraph{Multi-State and Hierarchical Memory.}
To improve the long-context capabilities of single-state linear attention and recurrent memory models, several methods partition the context into segments
and compress each segment in its own memory state. As a result, these models maintain a bank of states over the context. Memory Caching~\citep{behrouz2026memory} segments the context uniformly at fixed token length per segment.  Log-linear attention organizes the recurrent states hierarchically by growing a logarithmic number of states along a Fenwick-tree partition~\citep{guo2026log} of the sequence. Dynamic linear attention determines state boundaries based on token-level information variation while bounding the cache by merging low-information states~\citep{wang2026dynamic}. Segment-recurrent and compressive Transformers cache or compress past activations across chunks~\citep{dai2019transformer,rae2019compressive,bulatov2022recurrent,wu2026addressable}. These works primarily focus on how states are formed via content segmentation (\ie \textit{separation}). However, we identify how the information extracted from each state is merged (\ie \textit{aggregation}) as a key bottleneck in multi-state linear attention and propose a method that improves that derivation of aggregation weights.  

\paragraph{Routing over Memories via Aggregation.}
Multi-state linear attention and recurrent memory models extract information from each state and aggregate them to generate the final output representation of a token. Routing occurs during aggregation by adjusting the contribution of each segment to the final output. Mixture-of-Memories routes each token to one of several linear-attention states~\citep{du2026mom}. Sparse State Expansion routes tokens into state partitions with top-$k$ hard selection~\citep{pan2025scaling}. Gated Slot Attention reads a bounded set of slots through a softmax~\citep{zhang2024gated}.
Gated Residual Memory scores cached memory modules with a context-aware gate, with a sparse variant that selects the most relevant ones~\citep{behrouz2026memory}. Such routers descend from mixture-of-experts gating~\citep{shazeer2017outrageously,fedus2021switch} and from content-based addressing of an external memory~\citep{graves2014neural,sukhbaatar2015end,wu2022memorizing}. These aggregation methods assign weights based on position, the magnitude of its readout, or a single pooled key. However, \ours keeps first and second moments of each segment's keys to derive more informative aggregation weights.

\paragraph{Associative Recall and Long-Context Retrieval.}
Retrieving specific information from a long sequence is a core capability of models that handle long-context. In particular, such tasks evaluate the memory capacity of linear attention and recurrent memory models that compress the entire context into a single memory state. Multi-query associative recall and the Zoology analysis attribute much of the quality gap between recurrent memory models and standard transformers to recall~\citep{arora2024zoology}, and the recall-throughput tradeoff frames recall as bounded by recurrent state size~\citep{arora2024simple}. Copying and factual-recall analyses make the capacity limit of the memory state more precise~\citep{jelassi2024repeat,nichani2025understanding}. Furthermore, studies show that long-context models demonstrate positional biases where they prefer information located in certain depth within the context, usually close to the end~\citep{hsieh2024ruler,yen2025helmet}. We use these benchmarks to demonstrate that \ours mitigates such biases by improving aggregation of segment-level outputs.

\section{Experiment Details}
\label{app:config}

We describe our experimental setup in detail. \cref{app:model-config} describes the model and method configurations, \cref{app:train-config} the mid-training hyperparameters, \cref{app:datasets} the datasets and benchmarks, and \cref{app:evaluation} the evaluation protocol. In \cref{app:pseudocode} we provide the pseudocode for \ours.

\subsection{Model Configurations}
\label{app:model-config}

\Cref{tab:model-config} lists the two backbones and the \ours hyperparameters. Gated DeltaNet-1.3B (21 layers, hidden size 2048) was pre-trained on 50B tokens of FineWeb-edu, and Mamba-2-780M (48 layers, hidden size 1536) on 300B tokens of the Pile. For both, \ours uses a segment size of $m{=}256$ and a moment dimension of $\rdim{=}128$.

\begin{table}[h]
\centering
\small
\caption{\textbf{Model Configurations.}}
\label{tab:model-config}
\begin{tabular}{l c c}
\toprule
 & \textbf{Gated DeltaNet-1.3B} & \textbf{Mamba-2-780M} \\
\midrule
Parameters              & $\approx$1.3B      & $\approx$780M \\
Pretraining corpus      & FineWeb-edu        & Pile \\
Pretraining tokens      & 50B                & 300B \\
\midrule
Hidden size             & 2048               & 1536 \\
Layers                  & 21                 & 48 \\
Heads                   & 6                  & 48 \\
Head dimension          & 256                & 64 \\
SSM state size          & --                 & 128 \\
Expansion factor        & 2                  & 2 \\
Short-conv kernel       & 4                  & 4 \\
\midrule
\multicolumn{3}{@{}l}{\textit{\ours}} \\
Segment size $m$                         & 256 & 256 \\
Moment dimension $\rdim$                & 128 & 128 \\
Key/query normalization                  & \multicolumn{2}{c}{$\ell_2$-normalized} \\
Temperatures $\taum,\tauv$, gate $\gate$ & \multicolumn{2}{c}{learned} \\
\bottomrule
\end{tabular}
\end{table}

\subsection{Mid-Training Hyperparameters}
\label{app:train-config}

\Cref{tab:train-config} shows the mid-training recipe, which is shared by both backbones and by all m. We use AdamW with a weight decay of 0.1 and a peak learning rate of $1\times10^{-4}$, which follows a cosine schedule after 305 warmup steps. Each run uses 16{,}384-token sequences and a global batch of 131{,}072 tokens for 15{,}259 steps, which amounts to 2.0B tokens in bfloat16 mixed precision.

\begin{table}[h]
\centering
\small
\caption{\textbf{Mid-Training Hyperparameters.}}
\label{tab:train-config}
\begin{tabular}{l c}
\toprule
\textbf{Hyperparameter} & \textbf{Value} \\
\midrule
Optimizer                 & AdamW ($\epsilon=10^{-15}$) \\
Weight decay              & 0.1 \\
Peak learning rate        & $1\times10^{-4}$ \\
Learning-rate schedule    & cosine (min ratio $0.1$) \\
Warmup steps              & 305 \\
Gradient clipping (max norm) & 1.0 \\
Sequence length           & 16{,}384 \\
Global batch size         & 131{,}072 tokens \\
Training steps            & 15{,}259 \\
Total tokens              & 2.0B \\
Precision                 & bfloat16 (mixed) \\
Random seed               & 42 \\
Data                      & Long-Data-Collections, augmented with RULER NIAH/QA \\
\bottomrule
\end{tabular}
\end{table}

\subsection{Datasets and Benchmarks}
\label{app:datasets}

\Cref{tab:eval-summary} summarizes the mid-training data and the evaluation benchmarks, together with the metric, context length, and number of samples for each. We evaluate long-context retrieval on RULER~\citep{hsieh2024ruler} and long-context understanding on LongBench~\citep{bai2024longbench}, both at up to a 16K context.

\begin{table}[h]
\centering
\small
\caption{\textbf{Summary of Datasets and Evaluation Benchmarks.} We mid-train on Long-Data-Collections augmented with RULER-style NIAH/QA, and evaluate retrieval on RULER and long-context understanding on LongBench (filtered to $\le$16K tokens). Counts vary slightly by backbone tokenizer.}
\label{tab:eval-summary}
\begin{tabular}{l l l l l}
\toprule
\textbf{Phase} & \textbf{Dataset / Task} & \textbf{Metric} & \textbf{Context} & \textbf{Size} \\
\midrule
Mid-training & Long-Data-Collections ($+$ RULER NIAH/QA) & -- & 16K & 2.0B tokens \\
\midrule
\multirow{6}{*}{\shortstack[l]{RULER\\(retrieval)}}
& S-NIAH-1/2/3 & recall & 16K & 500 each \\
& MK-NIAH-1 & recall & 16K & 500 \\
& MQ-NIAH & recall & 16K & 500 \\
& MV-NIAH & recall & 16K & 500 \\
& HotpotQA & recall & 16K & 500 \\
& SQuADQA & recall & 16K & 500 \\
\midrule
\multirow{11}{*}{\shortstack[l]{LongBench}}
& NarrativeQA (NQ) & F1 & $\le$16K & 61 \\
& Qasper (QR) & F1 & $\le$16K & 197 \\
& MultiFieldQA (MF) & F1 & $\le$16K & 150 \\
& HotpotQA (HP) & F1 & $\le$16K & 161 \\
& 2WikiMQA (2W) & F1 & $\le$16K & 198 \\
& GovReport (GR) & ROUGE & $\le$16K & 174 \\
& QMSum (QM) & ROUGE & $\le$16K & 116 \\
& MultiNews (MN) & ROUGE & $\le$16K & 200 \\
& SAMSum (SS) & ROUGE & $\le$16K & 172 \\
& TREC (TC) & accuracy & $\le$16K & 200 \\
& TriviaQA (TQ) & F1 & $\le$16K & 142 \\
\bottomrule
\end{tabular}
\end{table}

\subsection{Evaluation Protocol}
\label{app:evaluation}

All evaluations use a 16K context length. We evaluate on RULER~\citep{hsieh2024ruler} and LongBench~\citep{bai2024longbench} following the official metrics and configurations from Language Model Evaluation Harness~\citep{eval-harness}. For RULER, we run the eight NIAH and QA tasks listed in \cref{tab:eval-summary} and report the score on each task together with the average across tasks. For LongBench, we run the eleven subtasks in \cref{tab:eval-summary}, keeping only the samples of at most 16K tokens, and score each task with its standard metric (F1, ROUGE, or accuracy).

For HELMET~\citep{yen2025helmet}, we use the TriviaQA retrieval-augmented generation task and follow the official evaluation setup. Each prompt contains several passages, and only one of them, the gold passage, contains the answer to the question. We randomly shuffle the position of the gold passage within the context and report accuracy as a function of that position, from the first sixth ($1/6$) to the last sixth ($6/6$) of the context.

\subsection{Pseudocode}
\label{app:pseudocode}

\begin{algorithm}[h]
\caption{\textbf{\ours{}} (Gated DeltaNet backbone) with equal-size segments $C_1,\dots,C_\nseg$ of $m$ tokens, learnable temperatures $\taum,\tauv$, and a learnable gate $\gamma_t$.}
\label{alg:merlin}
\begin{algorithmic}[1]
\For{$i = 1,\dots,\nseg$} 
    \State $\mS_{C_i},\ \firstm,\ \secondm \gets \mathbf{0},\ \mathbf{0},\ \mathbf{0}$ 
    \For{$t \in \seg$} 
        \State $\vq_t,\vk_t,\vv_t \gets \mW_q\vx_t,\ \mW_k\vx_t,\ \mW_v\vx_t$ 
        \State $\vq'_t,\vk'_t \gets \mW_{q'}\vx_t,\ \mW_{k'}\vx_t$ 
        \State $\mS_{C_i} \gets \mS_{C_i}\,(\alpha_t\,(\mI - \beta_t\vk_t\vk_t^\top)) + \beta_t\vv_t\vk_t^\top$ \Comment{GDN memory state, \cref{eq:gdn}}
        \State $\firstm \gets \firstm + \tfrac{1}{m}\vk'_t$,\quad $\secondm \gets \secondm + \tfrac{1}{m}\vk'_t{\vk'_t}^\top$ \Comment{moments of keys}
        \For{$\ell = 1,\dots,i$} 
            \State $\vo_{t,\ell} \gets \mS_{C_\ell}\vq_t$,\quad $r^\mu_{t,\ell} \gets \bm{\mu_\ell}^{\!\top}\vq'_t$,\quad $r^\Sigma_{t,\ell} \gets {\vq'_t}^\top\bm{\Sigma_\ell}\,\vq'_t - (r^\mu_{t,\ell})^2$
        \EndFor
        \State $\lambda_{t,\ell} \gets \gamma_t\,\softmax_\ell(\taum\,r^\mu_{t,\ell}) + (1-\gamma_t)\,\softmax_\ell(\tauv\,r^\Sigma_{t,\ell})$ \Comment{\cref{eq:weight}}
        \State $\vo_t \gets \textstyle\sum_{\ell \le i}\lambda_{t,\ell}\,\vo_{t,\ell}$ \Comment{aggregate readouts, \cref{eq:aggregation}}
    \EndFor
\EndFor
\State \Return $\vo_1,\dots,\vo_\seqlen$
\end{algorithmic}
\end{algorithm}

\begin{algorithm}[h]
\caption{\textbf{\ours{}} (Mamba-2 backbone) with equal-size segments $C_1,\dots,C_\nseg$ of $m$ tokens, learnable temperatures $\taum,\tauv$, and a learnable gate $\gamma_t$. The state $\mS_{C_i}\in\mathbb{R}^{P\times N}$, and the cached moments are shared per group.}
\label{alg:merlin-mamba}
\begin{algorithmic}[1]
\For{$i = 1,\dots,\nseg$}
    \State $\mS_{C_i},\ \firstm,\ \secondm \gets \mathbf{0},\ \mathbf{0},\ \mathbf{0}$
    \For{$t \in \seg$}
        \State $\vv_t,\mB_t,\mD_t,\Delta_t \gets \mW_v\vx_t,\ \mW_B\vx_t,\ \mW_D\vx_t,\ \operatorname{softplus}(\mW_\Delta\vx_t)$
        \State $\vq'_t,\vk'_t \gets \mW_{q'}\vx_t,\ \mW_{k'}\vx_t$
        \State $\mS_{C_i} \gets \exp(\Delta_t A)\,\mS_{C_i} + \Delta_t\,\vv_t\mB_t^\top$ \Comment{Mamba-2 SSD state}
        \State $\firstm \gets \firstm + \tfrac{1}{m}\vk'_t$,\quad $\secondm \gets \secondm + \tfrac{1}{m}\vk'_t{\vk'_t}^\top$ \Comment{moments of keys}
        \For{$\ell = 1,\dots,i$}
            \State $\vo_{t,\ell} \gets \mS_{C_\ell}\mD_t$,\quad $r^\mu_{t,\ell} \gets \bm{\mu_\ell}^{\!\top}\vq'_t$,\quad $r^\Sigma_{t,\ell} \gets {\vq'_t}^\top\bm{\Sigma_\ell}\,\vq'_t - (r^\mu_{t,\ell})^2$
        \EndFor
        \State $\lambda_{t,\ell} \gets \gamma_t\,\softmax_\ell(\taum\,r^\mu_{t,\ell}) + (1-\gamma_t)\,\softmax_\ell(\tauv\,r^\Sigma_{t,\ell})$ \Comment{\cref{eq:weight}}
        \State $\vo_t \gets \textstyle\sum_{\ell \le i}\lambda_{t,\ell}\,\vo_{t,\ell}$ \Comment{aggregate readouts, \cref{eq:aggregation}}
    \EndFor
\EndFor
\State \Return $\vo_1,\dots,\vo_\seqlen$
\end{algorithmic}
\end{algorithm}

\Cref{alg:merlin} shows \ours{} on a Gated DeltaNet backbone, assuming equal-size segments of $m$ tokens each. The outer loop iterates over the $\nseg$ segments, and the inner loop iterates over the $m$ tokens within each segment. For each token, the algorithm first updates the memory state of its segment together with the first and second moments of the segment's keys, so both accumulate token by token. It then reads out from every visible segment, meaning all closed segments plus the current, partially built one. Finally, it aggregates these readouts using the moment-derived weights from \cref{eq:weight}.

\Cref{alg:merlin-mamba} shows \ours{} on a Mamba-2 backbone. The structure is identical to \cref{alg:merlin}, but differs only on the memory state recurrence. Each token's hidden state is projected to the SSM value $\vv_t$, input matrix $\mB_t$ (the write key), output matrix $\mD_t$ (the read query), and step size $\Delta_t$. The segment state $\mS_{C_i}\in\mathbb{R}^{P\times N}$ ($P$ the head dimension, $N$ the SSM state size) follows the Mamba-2 SSD recurrence with per-head scalar decay $\exp(\Delta_t A)$. In Mamba-2, the cached moments are shared per group. The aggregation step is identical to \cref{eq:weight}.

\section{Additional Analysis}
\label{app:additional-analysis}

\paragraph{Memory Cost.} \Cref{tab:memory} reports the recurrent memory retained to decode each token, relative to the single fixed-state (Vanilla) cache. All multi-state methods cache the $L=\seqlen/m$ segment states, which dominate the cost. Position adds no cached moments, and Meanpool adds a negligible first moment ($\le0.1\%$), while \ours additionally caches a $\rdim{\times}\rdim$ second moment. This extra cost is controlled by $\rdim$ and is at most $12.6\%$ for Gated DeltaNet and $4.2\%$ for Mamba-2 (whose moments are shared per group) at $\rdim{=}128$, shrinking quadratically as $\rdim$ decreases.

\begin{table}[h]
\centering
\small
\caption{\textbf{Relative Decode Memory Cache} ($\times$ the single fixed-state Vanilla cache) at a 16K context, by segment size $m$ (number of states $L{=}\seqlen/m$) and moment dimension $\rdim$. All multi-state methods cache the $L$ segment states. Position caches no moments, Meanpool caches a first moment ($\le0.1\%$), and \ours caches a $\rdim{\times}\rdim$ second moment. \ours's overhead is larger for Gated DeltaNet (per-head router) than Mamba-2 (per-group router).}
\label{tab:memory}
\begin{subtable}[t]{0.49\textwidth}
\centering
\caption{Gated DeltaNet-1.3B ($d_k{\times}d_v{=}256{\times}512$)}
\resizebox{\linewidth}{!}{%
\begin{tabular}{l cccc}
\toprule
 & \multicolumn{4}{c}{$m$ (states $L$)} \\
\cmidrule(lr){2-5}
Method & 512 (32) & 256 (64) & 128 (128) & 64 (256) \\
\midrule
Vanilla & $1\times$ & $1\times$ & $1\times$ & $1\times$ \\
Position & $32.0\times$ & $64.0\times$ & $128.0\times$ & $256.0\times$ \\
Meanpool & $32.0\times$ & $64.0\times$ & $128.0\times$ & $256.0\times$ \\
\midrule
\ours\,($\rdim{=}16$)  & $32.1\times$ & $64.1\times$ & $128.3\times$ & $256.5\times$ \\
\ours\,($\rdim{=}32$)  & $32.3\times$ & $64.5\times$ & $129.0\times$ & $258.1\times$ \\
\ours\,($\rdim{=}64$)  & $33.0\times$ & $66.0\times$ & $132.1\times$ & $264.1\times$ \\
\ours\,($\rdim{=}128$) & $36.0\times$ & $72.1\times$ & $144.1\times$ & $288.3\times$ \\
\bottomrule
\end{tabular}}
\end{subtable}
\hfill
\begin{subtable}[t]{0.49\textwidth}
\centering
\caption{Mamba-2-780M ($P{\times}N{=}64{\times}128$)}
\resizebox{\linewidth}{!}{%
\begin{tabular}{l cccc}
\toprule
 & \multicolumn{4}{c}{$m$ (states $L$)} \\
\cmidrule(lr){2-5}
Method & 512 (32) & 256 (64) & 128 (128) & 64 (256) \\
\midrule
Vanilla & $1\times$ & $1\times$ & $1\times$ & $1\times$ \\
Position & $32.0\times$ & $64.0\times$ & $128.0\times$ & $256.0\times$ \\
Meanpool & $32.0\times$ & $64.0\times$ & $128.0\times$ & $256.0\times$ \\
\midrule
\ours\,($\rdim{=}16$)  & $32.0\times$ & $64.0\times$ & $128.1\times$ & $256.2\times$ \\
\ours\,($\rdim{=}32$)  & $32.1\times$ & $64.2\times$ & $128.3\times$ & $256.7\times$ \\
\ours\,($\rdim{=}64$)  & $32.3\times$ & $64.7\times$ & $129.4\times$ & $258.7\times$ \\
\ours\,($\rdim{=}128$) & $33.3\times$ & $66.7\times$ & $133.4\times$ & $266.7\times$ \\
\bottomrule
\end{tabular}}
\end{subtable}
\end{table}

\paragraph{Parameter Overhead.} We show the number of parameters that each aggregation method adds to the backbone under default model configurations ($m=256, \rdim=128$). The parameter overhead of \ours roughly equals the Meanpool baseline (4.5\% for Gated DeltaNet-1.3B and 2.2\% for Mamba-2-780M). The Position baseline is relatively lightweight, with only 1.2\% and 0.6\% overhead for Gated DeltaNet-1.3B and Mamba-2-780M respectively.

\begin{table}[h]
  \centering
  \small
  \caption{\textbf{Parameter Overhead of the Aggregation Mechanisms.} Parameters added in the aggregation pathway on top of each shared backbone. We assume default model configurations where $m=256, \rdim{=}128$.}
  \label{tab:params}
  \begin{tabular}{@{}l l r r@{}}
  \toprule
  Backbone & Aggregation & Added & \% \\
  \midrule
  \multirow{4}{*}{GatedDeltaNet-1.3B}
    & Vanilla           & $0$     & $0.0\%$ \\
    & \textsc{Position} & $16.8$M & $1.2\%$ \\
    & \textsc{Meanpool} & $66.1$M & $4.5\%$ \\
    & \ours{}           & $66.3$M & $4.6\%$ \\
  \midrule
  \multirow{4}{*}{Mamba-2-780M}
    & Vanilla           & $0$     & $0.0\%$ \\
    & \textsc{Position} & $4.8$M  & $0.6\%$ \\
    & \textsc{Meanpool} & $18.9$M & $2.2\%$ \\
    & \ours{}           & $18.9$M & $2.2\%$ \\
  \bottomrule
  \end{tabular}
  \end{table}

\paragraph{Qualitative Examples.} \Cref{tab:qualitative} illustrates the aggregation weight assignment on individual segments given retrieval-augmented generation samples in HELMET~\cite{yen2025helmet} (two Natural Questions, two TriviaQA, all at 16K context). For each query we show the gold segment that contains the ground truth as well as four randomly sampled distractor (irrelevant) segments, and the aggregation weight $\aggw$ each receives in the last layer. \ours assigns highest weights on the gold segment, well above the recency-driven \textsc{Position} router and above \textsc{Meanpool}. 

\begin{table*}[t]
\centering
\small
\caption{\textbf{Qualitative Examples.} We evaluate the GDN-based models on 4 random RAG samples (2 from Natural Questions and 2 from TriviaQA) in HELMET~\citep{yen2025helmet} 16K. We show the last layer's aggregation weights assigned to the ground truth and distractor segments for each aggregation method. \ours assigns highest weights to the ground truth segment compared to the Meanpool and Position baselines.}
\label{tab:qualitative}
\begin{tabular}{@{}p{0.05\textwidth} l p{0.36\textwidth} ccc@{}}
\toprule
Index & Type & Candidate segment & \textsc{Position} & \textsc{Meanpool} & \ours{} \\
\midrule
\rowcolor{qarow}\multicolumn{6}{@{}p{0.95\textwidth}@{}}{\textbf{Question:} ``what is another name for the water cycle?''} \\
\rowcolor{qarow}\multicolumn{6}{@{}p{0.95\textwidth}@{}}{\textbf{Answer:} hydrological cycle} \\
\addlinespace[2pt]
9 & Distractor & \dots of the world's water supply, about 321,000,000 cubic miles of it\dots & 0.006 & 0.018 & 0.017 \\
22 & Distractor & \dots water vapour or aqueous vapor is the gaseous phase of water\dots & 0.005 & 0.019 & 0.016 \\
\rowcolor{goldrow} \textbf{33} & \textbf{Ground Truth} & \textit{\dots also known as the hydrologic cycle or the \textbf{hydrological cycle}, the continuous movement of water\dots} & 0.005 & 0.030 & \cellcolor{hlcell}\textbf{0.055} \\
35 & Distractor & \dots cultural eutrophication of lakes is primarily due to phosphorus\dots & 0.005 & 0.008 & 0.005 \\
54 & Distractor & \dots loch or fen; each has a special meaning, though modern usage\dots & 0.005 & 0.010 & 0.007 \\
\midrule
\rowcolor{qarow}\multicolumn{6}{@{}p{0.95\textwidth}@{}}{\textbf{Question:} ``where are the Washington Redskins based out of?''} \\
\rowcolor{qarow}\multicolumn{6}{@{}p{0.95\textwidth}@{}}{\textbf{Answer:} the Washington metropolitan area} \\
\addlinespace[2pt]
8 & Distractor & \dots Franchise history: the Daniel Snyder ownership era (1999--present)\dots & 0.004 & 0.010 & 0.012 \\
28 & Distractor & \dots Daniel Snyder \dots W. Smith, the founder of Federal Express\dots & 0.004 & 0.016 & 0.013 \\
\rowcolor{goldrow} \textbf{37} & \textbf{Ground Truth} & \textit{\dots a professional football team based in \textbf{the Washington metropolitan area}; they compete in the NFL\dots} & 0.004 & 0.034 & \cellcolor{hlcell}\textbf{0.050} \\
39 & Distractor & \dots History. The Redskins relocated from Boston, Massachusetts\dots & 0.004 & 0.018 & 0.031 \\
50 & Distractor & \dots a home for that team; its highest-ranking teams as of 2015\dots & 0.004 & 0.030 & 0.027 \\
\midrule
\rowcolor{qarow}\multicolumn{6}{@{}p{0.95\textwidth}@{}}{\textbf{Question:} ``in which Charles Dickens novel does the character Mr M Choakumchild appear?''} \\
\rowcolor{qarow}\multicolumn{6}{@{}p{0.95\textwidth}@{}}{\textbf{Answer:} Hard Times} \\
\addlinespace[2pt]
16 & Distractor & \dots the illiterate poor chipped in ha'pennies to have each new monthly episode\dots & 0.004 & 0.010 & 0.024 \\
\rowcolor{goldrow} \textbf{26} & \textbf{Ground Truth} & \textit{\dots where he wrote ``Bleak House'' (1852--53) and \textbf{``Hard Times''} (1854)\dots} & 0.004 & 0.021 & \cellcolor{hlcell}\textbf{0.060} \\
42 & Distractor & \dots Mrs Nickleby is based on his mother, though she did not recognize herself\dots & 0.004 & 0.012 & 0.010 \\
45 & Distractor & \dots the novel ``Nicholas Nickleby'', and Mrs Micawber in ``David Copperfield''\dots & 0.004 & 0.023 & 0.010 \\
53 & Distractor & \dots Dickens in films and TV: Lennox Pawle (1935)\dots & 0.004 & 0.019 & 0.007 \\
\midrule
\rowcolor{qarow}\multicolumn{6}{@{}p{0.95\textwidth}@{}}{\textbf{Question:} ``which chain of European retailers began near a crossroads in Annecy, France in 1958?''} \\
\rowcolor{qarow}\multicolumn{6}{@{}p{0.95\textwidth}@{}}{\textbf{Answer:} Carrefour} \\
\addlinespace[2pt]
6 & Distractor & \dots the family also holds a stake in 3 Suisses, Norauto and others\dots & 0.007 & 0.021 & 0.016 \\
\rowcolor{goldrow} \textbf{25} & \textbf{Ground Truth} & \textit{\dots \textbf{Carrefour} S.A. is a French multinational retailer\dots} & 0.007 & 0.019 & \cellcolor{hlcell}\textbf{0.049} \\
31 & Distractor & \dots within its metropolitan boundaries; the Cagliari--Sarroch port system\dots & 0.007 & 0.013 & 0.016 \\
37 & Distractor & \dots 17 September 2012 \dots Jacqueline de Jong; an internment camp\dots & 0.007 & 0.008 & 0.006 \\
42 & Distractor & \dots some trades are also present along the avenue de Verdun\dots & 0.007 & 0.011 & 0.007 \\
\bottomrule
\end{tabular}
\end{table*}

\section{Limitations}

Due to compute constraints, our experiments cover two backbones (Gated DeltaNet-1.3B and Mamba-2-780M) at a 16K context length, and we mid-train multi-state models from single-state checkpoints. We compare \ours against Vanilla, Position, and Meanpool aggregation with uniform segmentation. Studying larger models, longer contexts, and a broader set of aggregation baselines is beyond our current compute budget, and we leave it to future work.

\end{document}